\documentclass[acmsmall,screen]{acmart}

\AtBeginDocument{%
  }

\setcopyright{acmcopyright}
\copyrightyear{xxx}
\acmYear{xxx}
\acmDOI{XXXXXXX.XXXXXXX}

\acmJournal{TKDD}

\usepackage{amsmath}
\usepackage{color}
\usepackage{multirow}

\usepackage{bbm}
\newtheorem{definition}{Definition}
\usepackage[noend]{algpseudocode}

\usepackage{amssymb}
\usepackage{hyperref}
\usepackage{algorithmicx,algorithm}
\usepackage{algpseudocode}

\begin{document}

\title{Boosting Fair Classifier Generalization through Adaptive Priority Reweighing}

\author{Zhihao Hu}
\authornote{These authors contributed to the work equally.}
\affiliation{
  \institution{University of Science and Technology of China}
  \city{Hefei}
  \state{Anhui}
  \postcode{230027}
  \country{China}
}
\email{huzhihao12@gmail.com}

\author{Yiran~Xu}
\authornotemark[1]
\affiliation{
  \institution{Warwick Business School, University of Warwick}
  \city{Coventry}
  \state{West Midlands}
  \postcode{CV4 7AL}
  \country{United Kingdom}
}
\email{Yiran.Xu@warwick.ac.uk}

\author{Mengnan Du}
\affiliation{
  \institution{New Jersey Institute of Technology}
  \city{Newark}
  \state{NJ}
  \postcode{07102}
  \country{United States}
}
\email{mengnan.du@njit.edu}

\author{Jindong Gu}
\affiliation{
  \institution{University of Oxford}
  \city{Oxford}
  \postcode{OX1 3PJ}
  \country{United Kingdom}
}
\email{jindong.gu@outlook.com}

\author{Xinmei Tian}
\authornote{Corresponding author.}
\affiliation{
  \institution{University of Science and Technology of China}
  \city{Hefei}
  \state{Anhui}
  \postcode{230027}
  \country{China}
}
\email{xinmei@ustc.edu.cn}

\author{Fengxiang He}
\affiliation{
  \institution{University of Edinburgh}
  \city{Edinburgh}
  \postcode{EH8 9AB}
  \country{United Kingdom}
}
\email{F.He@ed.ac.uk}

\renewcommand{\shortauthors}{Hu et al.}

\begin{abstract}
With the increasing penetration of machine learning applications in critical decision-making areas, calls for algorithmic fairness are more prominent. Although there have been various modalities to improve algorithmic fairness through learning with fairness constraints, their performance does not generalize well in the test set. A performance-promising fair algorithm with better generalizability is needed. This paper proposes a novel adaptive reweighing method to eliminate the impact of the distribution shifts between training and test data on model generalizability. Most previous reweighing methods propose to assign a unified weight for each (sub)group. Rather, our method granularly models the distance from the sample predictions to the decision boundary. Our adaptive reweighing method prioritizes samples closer to the decision boundary and assigns a higher weight to improve the generalizability of fair classifiers. Extensive experiments are performed to validate the generalizability of our adaptive priority reweighing method for accuracy and fairness measures (i.e., equal opportunity, equalized odds, and demographic parity) in tabular benchmarks. We also highlight the performance of our method in improving the fairness of language and vision models. The code is available at https://github.com/che2198/APW.
\end{abstract}




\received{14 September 2023}
\received[revised]{12 February 2024}
\received[accepted]{14 May 2024}

\maketitle

\section{Introduction}
With the increasing penetration of machine learning applications in critical decision-making areas, such as mortgage approval~\cite{madras2018learning}, employee selection~\cite{pessach2023algorithmic}, recidivism prediction~\cite{beutel2019fairness}, and healthcare, calls for algorithmic fairness are more prominent. In reality, machine learning tools can behave in unintentional or even harmful ways~\cite{wan2023processing}. For example, research on the COMPAS dataset~\cite{chouldechova2017fair} shows that a well-calibrated classification algorithm tends to classify African-American defendants with a higher risk of re-conviction while classifying white defendants with lower risk~\cite{angwin2022machine}. In this case, the machine learning algorithms perpetuated and amplified the long-standing discrimination against people of color~\cite{wang2019balanced}. Both categorizations produced during data collection and/or analysis and the data itself are persistent, which paves the way for algorithms to compound the impact of historically biased data. More importantly, bias exists in forms of distribution, thus almost impossible to address bias by discarding individual instances. Therefore, it is much more difficult to achieve fairness by removing sensitive attributes from the data. Machine learning researchers have been focusing on higher classification accuracy, yet our study suggests the need to shift away from the fixation on accuracy and be actively attentive to mitigating biases from training data and ensuring generalizable performance across real-world datasets.

\begin{figure*}
    \centering
    \includegraphics[width=1\linewidth]{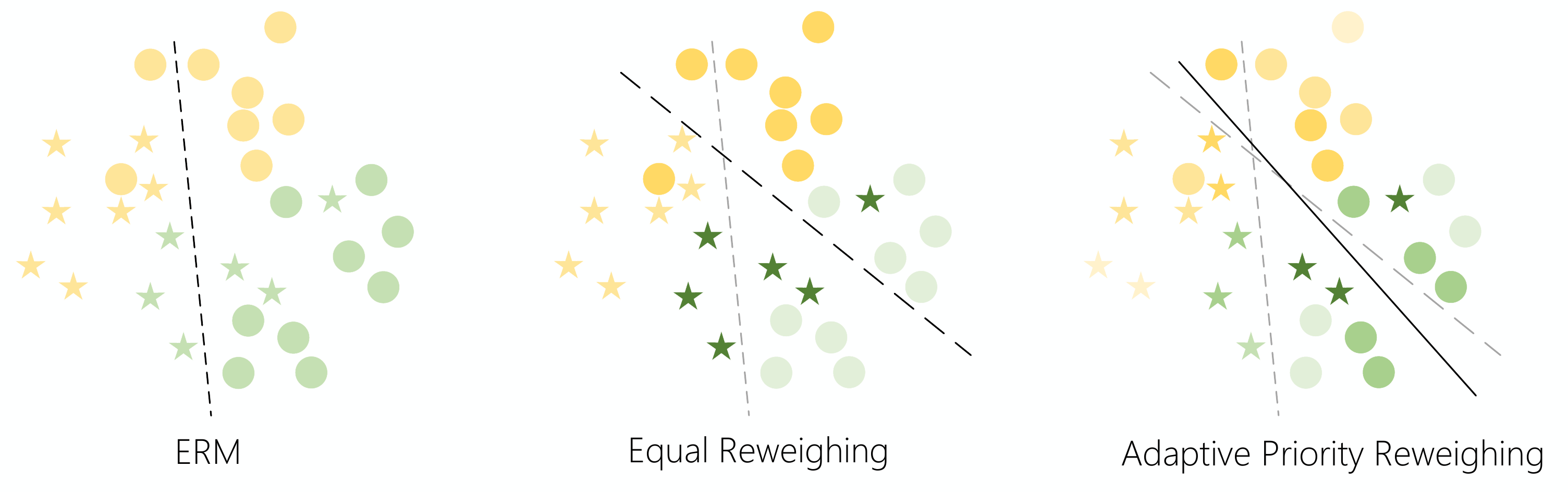}
    \caption{Illustration of our adaptive priority reweighing method. Samples of different shapes indicate samples with different labels, different colors refer to different sensitive attributes, and darker color indicates higher weight. The dotted lines represent the decision boundaries for Empirical Risk Minimization(ERM) and Equal Reweighing, while the full line represents the decision boundary for Adaptive Priority Reweighing. In ERM, fairness is disregarded altogether; all samples have the same weight, which could easily lead to unfairness. In equal reweighing, (sub)groups are created and all points within one (sub)group are assigned the same weight, ensuring improved fairness on the training set. However, using equal reweighing methods induces many points clustered around the decision boundary, which limits the generalization of fair classifiers. In comparison, our method models the distance from the sample to the decision boundary, which improves the generalizability of fair classifiers.}
\label{fig:method}
\end{figure*}

\begin{figure}
    \centering
    \includegraphics[width=0.8\linewidth]{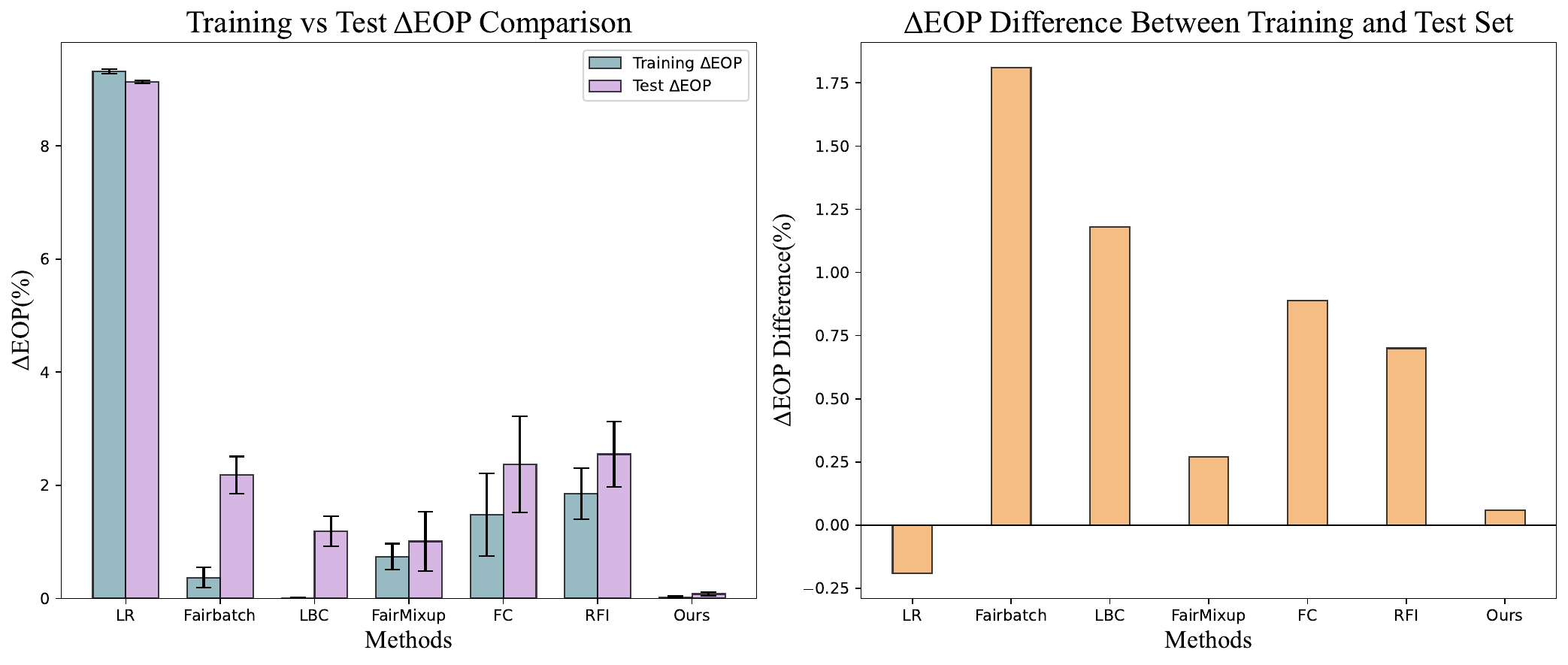}
    \caption{Comparison of Training and Testing of Equal Opportunity Gap~\cite{hardt2016equality}($\Delta_{\mathrm{EOP}}$). We first use a histogram to illustrate the performance of six methods on the Equal Opportunity fairness metric on the training and test set, showcasing the range of $\Delta_{\mathrm{EOP}}$ using error bars. In the histogram on the right, the difference in model performance is measured by subtracting the performance metrics on the training sets from those on the test set. The generalizability issue for fairness measures becomes particularly salient for algorithms with commendable fairness performance, such as Fairbatch~\cite{roh2020fairbatch} and Label Bias Correction(LBC)~\cite{jiang2020identifying}. For example, these algorithms achieve considerable fairness performance on the training set. However, the same level of fairness performance does not necessarily translate to the test set. Our method demonstrates superior efficacy in ensuring fairness performance on both training and test sets, achieving optimal generalizability.}
\label{fig:compare}
\end{figure}

In the recent decade, machine learning researchers have engaged in the conversation of algorithmic fairness by investigating the correlation between model output and sensitive attributes (e.g., gender and race). Diverse fairness metrics are proposed to measure the disparities in model output between different sensitive attributes. In order to improve model performance in measuring fairness, we identify fairness as a constraint term, enabling fairness to be improved by solving the constrained optimization problem. Existing methods use various regularization terms or constraints to address the problem, ranging from covariance~\cite{zafar2017fairness}, equalized odds~\cite{donini2018empirical},  to Rényi correlation~\cite{baharlouei2019renyi}. Although they provide a shared foundation for machine learning researchers to engage in the fairness conundrum, they all suffer from generalization issues. That is, although these methods minimize the value of constraints on the training set, they cannot always guarantee fairness on the test set (see Figure \ref{fig:compare}). Chuang and Mroueh (2021)~\cite{chuang2021fair} and Ramachandran and Rattani (2022)~\cite{ramachandran2022deep} address the generalization issues of algorithmic fairness using data augmentation. Chuang and Mroueh (2021) propose applying mixup path regularization to enhance the generalizability of the fair classifier. Ramachandran and Rattani (2022) further build on the conversation by engaging with the generative model to mitigate gender classification bias. Unfortunately, both studies require training data to conform to a specific distribution, leading to a limited representation of real-world datasets and reduced accuracy.

We aim to address the generalization issue of fairness improvement, but at the same time preserve accuracy. To do that, we propose an adaptive priority weight assignment strategy to improve fairness. As illustrated in Figure 1, we first divide the subgroups using sensitive attributes and predictions of the classifier. Data points closer to the decision boundary are more likely to be wrongly classified~\cite{kamiran2012data} — a phenomenon highlighting the critical role of decision boundaries in a model's generalizability~\cite{almeida2021mitigating}. Therefore, our method models the distance from data points to the decision boundary to improve the generalizability of fair classifiers. This solves the generalizability issues associated with equal reweighing. Equal reweighing methods induce the cluster of sample points around the decision boundary. Distribution shifts could cause the model to classify sample points closer to the decision boundary incorrectly, thus inducing unfairness in the test set and limiting the generalization of fair classifiers. 
Additionally, we take the statistical independence of model output value and sensitive attribute as the termination condition and assign higher weights to samples whose model output is closer to the decision boundary in each subgroup. 

We design experiments to validate the generalizability of our adaptive priority reweighing method for accuracy and fairness measures (i.e., equal opportunity, equalized odds, and demographic parity) on tabular, language, and vision benchmarks. We further highlight the performance of our method in improving the fairness of language and vision models. Our method shows promising results in improving the fairness of any pre-trained models simply via fine-tuning. The code is available at https://github.com/che2198/APW.

We summarize our contributions as follows:

1. We propose a novel reweighing method to mitigate data bias under different fairness
measures, whose performance is validated through experiments on tabular, language, and vision benchmarks.  

2. Our method provides a theoretical guarantee of generalization error bound.

3. Our method improves the fairness of any pre-trained model via fine-tuning by experimenting with both language and vision benchmarks.

\section{Related Work}
\textbf{Machine Learning Fairness.}
Most fairness-promoting algorithms focus on binary classification under binary-sensitive attributes. Generally, these methods can be divided into three categories: pre-processing, in-processing, and post-processing. Pre-processing methods tend to alter the sample distributions of protected variables to remove discrimination from the training data. Approaches to preprocessing data include sample selection~\cite{roh2020fairbatch, roh2021sample, zhang2020fairness}, sample reweighing~\cite{kamiran2012data, jiang2020identifying, chai2022fairness}, fair representation learning~\cite{zemel2013learning, madras2018learning}, fair data generation~\cite{xu2018fairgan, xu2019achieving, jang2021constructing}, etc. In-processing methods incorporate changes into the objective function or impose a constraint to remove discrimination during the model training process. Fairness constraints measure the correlation between model outputs and sensitive attributes~\cite{berk2017convex, aghaei2019learning}, such as adversarial layers~\cite{zhang2018mitigating, adel2019one},  contrastive learning loss~\cite{zhou2021contrastive}, uncorrelation constraints~\cite{zafar2017fairness}, counterfactual logit pairing~\cite{garcia2021maxmin}, and mutual information~\cite{baharlouei2019renyi, roh2020fr}. Post-processing transforms model output in different demographic groups to achieve fairness~\cite{hardt2016equality, fish2016confidence, woodworth2017learning, pleiss2017fairness}. These methods can satisfy the fairness constraints on the training set but may not be generalized to the test set.

Recently, some methods based on data augmentation have been proposed to enhance the generalizability of fair classifiers, such as mix-up paths regularization~\cite{chuang2021fair}, data generation~\cite{ramachandran2022deep}. However, these methods require training data to conform to a specific distribution or lead to computationally expensive algorithms. 

\noindent\textbf{Data Reweighing.} The concept of data reweighing has gained considerable attention as an effective approach to improving the performance and fairness of machine learning classifiers. Various studies have explored different aspects of data reweighing techniques to address the challenges associated with biased training data and classifier fairness. 

Many approaches introduced data reweighing as a means of improving classifier generalization~\cite{chu2022dna,zhou2022model}. They highlighted the importance of reweighing samples in the training dataset to mitigate issues arising from class imbalance, leading to improved overall classifier performance. Meanwhile,~\cite{liu2015classification,zhang2021deep} demonstrated that reweighing techniques can be instrumental in reducing bias during training, and then promoting the fairness of classifiers across diverse demographic groups.

A subset of research has concentrated specifically on fairness improvement through data reweighing. Kamiran et al.~\cite{kamiran2012data} pioneered the concept of fixed reweighing, which involves assigning different weights to instances from underrepresented groups to balance their influence during training. Adaptive reweighing methods, as proposed by~\cite{krasanakis2018adaptive,li2022achieving,wen2022distributionally,jiang2020identifying}, dynamically adjust instance weights based on the evolving characteristics of the dataset. These methods attempt to alleviate bias and enhance fairness during training. 

However, these data reweighing methods on fairness minimize constraints on the training set. They cannot guarantee fairness on the test set. Though Chai and Wang~\cite{chai2022fairness} attempt to address this by assigning varying weights to individual samples within each subgroup, it may not adequately address the generalizability issue, for it only allocates higher weight on sample point with higher loss. Therefore, we propose a novel reweighing method that addresses the generalization issue of fairness improvement and preserves classification accuracy by allocating higher weight to sample points closer to the decision boundary.

\section{Preliminaries}
We consider the binary classification task.
A sample is $z=(x,a,y)$, where $x \in \mathcal{X}$ is the input feature, $a \in \mathcal A = \{0,1\}$ is the sensitive attribute ({\it e.g.}, gender, race, and age), and $y \in \{0,1\}$ is the prediction target.
Let $Z$, $X$, $A$, and $Y$ denote the corresponding random variables of $z$, $x$, $a$, and $y$, respectively.
Then, the goal of fair machine learning is to learn a binary classifier $h_{\theta}: \mathcal{X} \times \mathcal A  \rightarrow [0,1]$ while ensuring a specific notion of fairness with respect to the sensitive attribute $A$, where $\theta$ is the parameter of the classifier.
For simplicity, we denote $\hat Y := f(h_{\theta}(X,A)) \in \{0,1\}$ as the prediction of the classifier $h$ for variable $Z=(X,A,Y)$, where $f$ is a label function.

\subsection{Group fairness}
A major family of fairness concepts is the {\it group fairness}, which aims to characterize discrimination across various groups of individuals.
The most used definitions of group fairness are {\it demographic parity}~\cite{feldman2015certifying}, {\it equalized odds}~\cite{hardt2016equality}, and {\it equal opportunity}~\cite{hardt2016equality}.

\begin{definition}[Fairness Definitions]
\label{def:fair_def}
Given a data distribution $p_z$, a classifier $h$ satisfies:
\begin{itemize}
\item Demographic parity if the prediction $\hat Y$ is independent of the sensitive attribute $A$.
\item Equalized odds if the prediction $\hat Y$ and the sensitive attribute $A$ are independent conditional on the target $Y$.
\item Equal opportunity if
$\mathbb{P}(\hat Y=1 | A=0, Y=1) = \mathbb{P}(\hat Y=1 | A=1, Y=1)$.
\end{itemize}
\end{definition}

The following notions are designed to assess the degree to which the classifier satisfies the fairness constraints presented in Definition \ref{def:fair_def}.

\begin{definition}[Fairness Measures]
\label{def:fair_measure}
Given a data distribution $p_z$, for a binary classifier $h$:
\begin{itemize}
\item
The demographic parity $\Delta_{\mathrm{DP}}$ is defined as:
\begin{align}
\Delta_{\mathrm{DP}}(\hat Y, p_z)
= | \mathbb{P}(\hat Y=1 | A=0) 
 - \mathbb{P}(\hat Y=1 | A=1) |.
\end{align}

\item
The equalized odds gap $\Delta_{\mathrm{EO}}$ is defined as:
\begin{align}
\Delta_{\mathrm{EO}}(\hat Y, p_z)&= \max_{y \in \{0,1\}} | \mathbb{P}(\hat Y=1 | A=0,Y=y) 
 - \mathbb{P}(\hat Y=1 | A=1,Y=y) |.
\end{align}

\item
The equal opportunity gap $\Delta_{\mathrm{EOP}}$ is defined as:
\begin{align}
\Delta_{\mathrm{EOP}}(\hat Y, p_z)
= | \mathbb{P}(\hat Y=1 | A=0, Y=1) 
 - \mathbb{P}(\hat Y=1 | A=1, Y=1) |.
\end{align}
\end{itemize}
\end{definition}

Combined with Definition \ref{def:fair_def}, it is clear that a small value of the fairness measure in Definition \ref{def:fair_measure} would demonstrate a strong non-discriminatory of the given classifier, vice versa.
When the notion $\Delta_{\mathrm{DP}}$, $\Delta_{\mathrm{EO}}$, or $\Delta_{\mathrm{EOP}}$ is equal to zero, the classifier $h$ perfectly satisfies demographic parity, equalized odds, or equal opportunity, respectively.

\subsection{Subgroup weights}
We divided the training samples into subgroups according to the sensitive attributes and predictions of the classifier. Specifically, we define three types of subgroups: $G_{y, a} := \left\{\left(x_i, a_i, y_i\right): y_{i} = y, a_{i} = a\right\}$, $G_{\ast, a} := \left\{\left(x_i, a_i, y_i\right): a_{i} = a\right\}$, and $G_{y, \ast} := \left\{\left(x_i, a_i, y_i\right): y_{i} = y\right\}$. Let \textit{m} represent the total number of training samples. We assigned the subgroup weight $W_{y, a}$ to samples that satisfy $Y = y$ and $A = a$. To quantify the number of samples within a subgroup, we introduced the notation $m_{y, a}$, defined as $m_{y, a}:= \left|\left\{i: y_{i} = y, a_{i} = a\right\}\right|$. Similarly, we defined $m_{y, \ast} := \left|\left\{i: y_{i} = y\right\}\right|$ and $m_{\ast, a} := \left|\left\{i: a_{i} = a\right\}\right|$. Additionally, we introduced the proportion $p_{y, a}$, which represents the fraction of samples in $G_{y, a}$ compared to the total number of samples: $p_{y, a}:= \frac{m_{y, a}}{m}$. For the equal reweighing approach, each sample is assigned a weight equal to its subgroup weight.

\subsection{Generalization error}
When dealing with the problem of an unknown distribution D, the empirical risk is commonly employed as an approximation of the expected risk. The empirical risk is defined as follows:
\begin{equation}
\hat{R}_{L, D}(h)=\frac{1}{m}\sum_{i=1}^{m}L\left(y_i, h\left(x_i, a_i\right)\right),
\end{equation}
where the corresponding expected risk is
\begin{equation}
R_{L, D}(h)=R[D, h, L]=E_{(X, A, Y) \sim D}\left[\hat{R}_{L, D}(h)\right],
\end{equation}
and $L$ represents a surrogate loss function, such as square loss, logistic loss, or hinge loss. The classifier is learned through the process of empirical risk minimization (ERM)~\cite{vapnik1999nature}:
\begin{equation}
h'=\arg \min _{h \in H} \hat{R}_{L, D}(h).
\end{equation}

The consistency of $\hat{R}_{L, D}(h')$ with respect to $\min _{h \in H} R_{L, D}(h)$ is of utmost importance when designing surrogate loss functions and learning algorithms. Let
\begin{equation}
h^*=\arg \min _{h \in H} R_{L, D}(h).
\end{equation}

It has been demonstrated in previous work~\cite{anthony1999neural} that the following proposition holds true:
\begin{equation}
R_{L, D}\left(h'\right)-R_{L, D}\left(h^*\right) \leq 2 \sup _{h \in H}\left|R_{L, D}(h)-\hat{R}_{L, D}(h)\right|.
\end{equation}

The term on the right-hand side is commonly referred to as the generalization error, which measures the performance of the learned model on unseen data. The consistency of the model is ensured through the convergence of the generalization error.

\section{Proposed Method}
In this section, we present our Adaptive Priority Reweighing method. First, we introduce the updating rules for adaptive priority reweighing, followed by its application to prominent fairness measures, including equal opportunity, equalized odds, and demographic parity. Finally, we provide the theoretical guarantee of our method.

\subsection{Updating Rules for Adaptive Priority Reweighing}
For a given dataset $\{\left(x_i, a_i, y_i\right)\}_{i=1}^m$ of $m$ samples, the vanilla training objective without reweighing can be formulated as:
\begin{equation}
    \mathop{\min}_{\theta}\frac{1}{m}\sum_{i=1}^{m}L\left(y_i, h_{\theta}\left(x_i, a_i\right)\right).
    \label{erm}
\end{equation}

Given that the unweighted training objective does not incorporate a fairness criterion, its minimizer often fails to satisfy the desired fairness requirement. To overcome this limitation, we introduce a weighting scheme for each sample to ensure the desired fairness guarantee. For example, assigning higher weights to a specific sensitive group can result in improved accuracy specifically for that group. Consequently, the batch gradient estimate provides an unbiased estimation of the reweighed empirical risk. In other words, if we draw a training example $i$ with weight $w_i$, the training objective can be formulated as follows:
\begin{equation}
    \mathop{\min}_{\theta}\frac{1}{m}\sum_{i=1}^{m}w_{i}L\left(y_i, h_{\theta}\left(x_i, a_i\right)\right).
\end{equation}

This observation leads to the following bilevel optimization-based interpretation of how reweighing interacts with the inner optimization algorithm. Initially, batch SGD optimizes the unweighted empirical risk. Subsequently, based on the results of the inner optimization, the outer optimizer iteratively refines the weights $\boldsymbol{w}:=\left(w_1, w_2, \ldots, w_m\right)$ assigned to each training example. The inner optimizer then operates on batches drawn from a new distribution, reoptimizing the inner objective function. This process continues until convergence is achieved. Consequently, the Adaptive Priority Reweighing algorithm can be perceived as a combination of an outer optimizer and an inner optimizer, solving the following bilevel optimization problem:
\begin{equation}
    \min _{\boldsymbol{w}} \operatorname{Cost}\left(\boldsymbol{\theta}_{\boldsymbol{w}}\right), 
\end{equation}
\begin{equation}
    \boldsymbol{\theta}_{\boldsymbol{w}}=\underset{\boldsymbol{\theta}}{\arg \min }\frac{1}{m}\sum_{i=1}^{m}w_{i}L\left(y_i, h_{\theta}\left(x_i, a_i\right)\right),
\end{equation}
where $\operatorname{Cost}\left(\cdot\right)$ captures the goal of the optimization. To simplify the discussion, we initially focus on the fairness metric $\Delta_{\mathrm{DP}}$, and then extend our analysis to include $\Delta_{\mathrm{EO}}$ and $\Delta_{\mathrm{EOP}}$. The outer optimization problem, specifically for the $\Delta_{\mathrm{DP}}$ metric, takes the following form:
\begin{align}
    \min _{\boldsymbol{w}}|\frac{1}{m_{1, \ast}}\sum_{z_i\in G_{1, \ast}}f(h_{\boldsymbol{\theta}_{\boldsymbol{w}}}\left(x_i, a_i\right)) - \frac{1}{m_{0, \ast}}\sum_{z_j\in G_{0, \ast}}f(h_{\boldsymbol{\theta}_{\boldsymbol{w}}}\left(x_j, a_j\right))|.
\end{align}

To address this optimization problem, we begin by computing the subgroup weight $G_{y, a}$ for each subgroup. Subsequently, we determine the sample weight based on the distance of each sample from the decision boundary within its respective subgroup. Now, let us delve into the rationale behind the calculation of the subgroup weight $G_{y, a}$.

If the classifier $h_{\theta}$ exhibits fairness, meaning that $A$ and $\hat Y$ are statistically independent, the expected probability $P_{\text{exp}}(\hat Y=1, A=1)$ can be expressed as follows:
\begin{align}
P_{\exp }(\hat Y=1, A=1):=\frac{\mid\{i: \hat{y}_{i} = 1\}\mid}{m} \times \frac{\mid\{i: a_{i} = 1\}\mid}{m}.
\end{align}

In reality, the observed probability is given by:
\begin{equation}
P_{obs }(\hat Y=1, A=1):=\frac{\mid\{i: \hat{y}_{i} = 1, a_{i} = 1\}\mid}{m},
\end{equation}
where $m$ represents the total number of samples. It is possible that the expected probability differs from the observed probability. If the expected probability exceeds the observed probability, it indicates a bias towards the subgroup $G_{1,1}$.

To address the bias, we aim to assign lower weights to objects that have been either deprived or favored. For this purpose, we assign weights to each subgroup $G_{y,a}$ as follows:
\begin{align}
W_{y,a}:=\frac{P_{\exp }(\hat Y=y, A=a)}{P_{obs }(\hat Y=y, A=a)} =\frac{\mid\{i:\hat{y}_{i} = y\}\mid \times \mid\{i: a_{i} = a\}\mid}{m \times \mid\{i: \hat{y}_{i} = y, a_{i} = a\}\mid}.
\end{align}

The weight assigned to each subgroup will be calculated as the ratio between the expected probability of observing an instance with its sensitive attribute value and prediction under the assumption of independence, and its corresponding observed probability.

Based on the aforementioned considerations, we propose the following algorithm for determining the subgroup weights:
\begin{align}
\forall t \in\{1,2, \ldots, T\}: W_{y,a}^{(t)}=W_{y,a}^{(t-1)}\times \frac{\mid\{i:\hat{y}_{i}^{(t-1)} = y\}\mid \times \mid\{i: a_{i} = a\}\mid + \alpha}{m \times \mid\{i: \hat{y}_{i}^{(t-1)} = y, a_{i} = a\}\mid + \alpha},
\end{align}
where $\alpha$ is a constant that regulates the magnitude of the weight adjustment, and $T$ represents the training iteration. The weight assigned to each sample is calculated as follows:
\begin{align}
\forall t \in\{1,2, \ldots, T\}: w_i^{(t)}=\frac{W_{y,a}^{(t)}}{\sum_{y} \sum_{a} W_{y,a}^{(t)}}\times \frac{p_{y,a}\exp(-\eta \phi_i^{(t)})}{\sum_{i=1}^{m_{y,a}}\exp(-\eta \phi_i^{(t)})}, z_i\in G_{y, a},
\end{align}
where $\eta$ represents the step size, and $\phi_i^{(t)}$ denotes the margin between the prediction of sample $z_i$ and the decision boundary value $d$:
\begin{align}
\forall t \in\{1,2, \ldots, T\}:\phi_i^{(t)} = |h_{\theta}^{(t)} - d|, i = 1, 2, \ldots, m,
\end{align}

For a comprehensive implementation of the fair classifier $h_{\theta}$ targeting Demographic Parity, please refer to Algorithm \ref{alg:dp}.

\begin{algorithm}
\caption{Training a fair classifier for Demographic Parity.}
\label{alg:dp}
\begin{algorithmic}[1]
\State {\bfseries Input:} Training dataset $\mathcal{Z}_{[m]}=\left\{\left(x_i, a_i, y_i\right)\right\}_{i=1}^m$, training iteration $T$, 
subgroup learning rate $\alpha$, step size $\eta$, value of decision boundary $d$, classification procedure $H$.
\State {\bfseries Output:} Fair classifier $h_{\theta}$ for Demographic Parity.
    \State Initialize the model parameters $\theta$ of the classifier $h$, subgroup weights $W_{1,1}^{(0)}=W_{1,0}^{(0)}=W_{0,1}^{(0)}=W_{0,0}^{(0)}=1$, sample weights $w_1^{(0)}=w_2^{(0)}=\ldots=w_m^{(0)}=\frac{1}{m}$, margin $\phi_1^{(0)}=\phi_2^{(0)}=\ldots=\phi_m^{(0)}=0$, respectively.
    \State Let $h_{\theta}^{(0)}:=H(\mathcal{Z}_{[m]},\{w_i^{(0)}\}_{i=1}^m)$.
    \For{$t$ \textbf{in} $1,\cdots,T$}
        \State Update the margin for each sample:
        \begin{gather*}
            \phi_i^{(t)} = |h_{\theta}^{(t)} - d| \text{ for } i \in [m].
        \end{gather*}
        \For{$y \in \{0, 1\}$}
            \For{$a \in \{0, 1\}$}
                \State Calculate subgroup weight $W_{y,a}^{(t)}$:
                \begin{align}
                    W_{y,a}^{(t)}=W_{y,a}^{(t-1)} \times \frac{\mid\{i:\hat{y}_{i}^{(t-1)} = y\}\mid \times \mid\{i: a_{i} = a\}\mid + \alpha}{m \times \mid\{i: \hat{y}_{i}^{(t-1)} = y, a_{i} = a\}\mid + \alpha}.\nonumber
                \end{align}
                \State Update sample weights:
                \begin{gather*}
                    w_i^{(t)}=\frac{W_{y,a}^{(t)}}{\sum_{y} \sum_{a} W_{y,a}^{(t)}}\times \frac{p_{y,a}\exp(-\eta \phi_i^{(t)})}{\sum_{i=1}^{m_{y,a}}\exp(-\eta \phi_i^{(t)})} \text{ for } z_i\in G_{y, a}.
                \end{gather*}
            \EndFor
        \EndFor
        \State Update $h_{\theta}^{(t)}:=H(\mathcal{Z}_{[m]},\{w_i^{(t)}\}_{i=1}^m)$.
    \EndFor \\
    \Return $h_{\theta}^{(T)}$
\end{algorithmic}
\end{algorithm}

Intuitively, if the observed positive prediction rate for a protected class $A = a$ is lower than the expected positive prediction rate, we adjust the weights of the positively labeled samples of $A = a$ by increasing them, while decreasing the weights of the negatively labeled samples of $A = a$. This approach encourages the classifier to improve its accuracy on the positively labeled samples within $A = a$, while reducing its accuracy on the negatively labeled samples. Moreover, the weight assigned to a sample is higher when it is closer to the decision boundary within each subgroup. These two mechanisms collectively contribute to an increase in the positive prediction rate for $A = a$ and enhance the generalizability of the fair classifier $h_{\theta}$.

Algorithm \ref{alg:dp} operates through a series of iterative steps as follows:
\begin{itemize}
    \item Update the margin for each sample.
    \item Adjust subgroup weights by the observed probability and the expected probability.
    \item Compute the weights for each sample based on these margin and subgroup weights.
    \item Retrain the classifier given these weights.
\end{itemize}

Algorithm \ref{alg:dp} takes as input a classification procedure $H$, which, given a dataset $\mathcal{Z}{[m]}=\left\{\left(x_i, a_i, y_i\right)\right\}{i=1}^m$ and weights ${w_i}_{i=1}^m$, outputs a classifier. In practice, $H$ can be any training procedure that minimizes a weighted loss function over a specific parameterized function class, such as logistic regression.

\subsection{Extensions to Other Fairness Notions}
\textbf{Equalized Odds:} Equalized odds requires that the prediction $\hat Y$ and the sensitive attribute $A$ are independent conditional on the target $Y$. Case 1, if the observed true positive rate for a protected class $A = a$ is lower than the expected true positive rate, we increase the weights of the positively labeled samples of $A = a$. This will encourage the classifier to increase its accuracy on the positively labeled samples in $A = a$. Case 2, if the observed true negative rate for a protected class $A = a$ is lower than the expected true negative rate, we increase the weights of the negatively labeled samples of $A = a$. This will encourage the classifier to increase its accuracy on the negatively labeled samples in $A = a$. In addition, the closer the sample is to the decision boundary in each subgroup, the higher the weight is assigned. See the full pseudocode of learning fair classifier $h_{\theta}$ for Equalized Odds in Algorithm\ref{alg:eo}.

\begin{algorithm}
\caption{Training a fair classifier for Equalized Odds.}
\label{alg:eo}
\begin{algorithmic}[1]
\State {\bfseries Input:} Training dataset $\mathcal{Z}_{[m]}=\left\{\left(x_i, a_i, y_i\right)\right\}_{i=1}^m$, training iteration $T$, 
subgroup learning rate $\alpha$, step size $\eta$, value of decision boundary $d$, classification procedure $H$.
\State {\bfseries Output:} Fair classifier $h_{\theta}$ for Equalized Odds.
    \State Initialize the model parameters $\theta$ of the classifier $h$, subgroup weights $W_{1,1}^{(0)}=W_{1,0}^{(0)}=W_{0,1}^{(0)}=W_{0,0}^{(0)}=1$, sample weights $w_1^{(0)}=w_2^{(0)}=\ldots=w_m^{(0)}=\frac{1}{m}$, margin $\phi_1^{(0)}=\phi_2^{(0)}=\ldots=\phi_m^{(0)}=0$, respectively.
    \State Let $h_{\theta}^{(0)}:=H(\mathcal{Z}_{[m]},\{w_i^{(0)}\}_{i=1}^m)$.
    \For{$t$ \textbf{in} $1,\cdots,T$}
        \State Update the margin for each sample:
        \begin{gather*}
            \phi_i^{(t)} = |h_{\theta}^{(t)} - d| \text{ for } i \in [m].
        \end{gather*}
        \For{$y \in \{0, 1\}$}
            \For{$a \in \{0, 1\}$}
                \State Calculate subgroup weight $W_{y,a}^{(t)}$:
                \begin{align}
                    W_{y,a}^{(t)}=W_{y,a}^{(t-1)} \times \frac{\mid\{i:\hat{y}_{i}^{(t-1)} = y, y_i = y\}\mid \times \mid\{i: y_i = y, a_{i} = a\}\mid + \alpha}{m_{y,\ast} \times \mid\{i: \hat{y}_{i}^{(t-1)} = y, y_i = y, a_{i} = a\}\mid + \alpha}.\nonumber
                \end{align}
                \State Update sample weights:
                \begin{gather*}
                    w_i^{(t)}=\frac{W_{y,a}^{(t)}}{\sum_{y} \sum_{a} W_{y,a}^{(t)}}\times \frac{p_{y,a}\exp(-\eta \phi_i^{(t)})}{\sum_{i=1}^{m_{y,a}}\exp(-\eta \phi_i^{(t)})} \text{ for } z_i\in G_{y, a}.
                \end{gather*}
            \EndFor
        \EndFor
        \State Update $h_{\theta}^{(t)}:=H(\mathcal{Z}_{[m]},\{w_i^{(t)}\}_{i=1}^m)$.
    \EndFor \\
    \Return $h_{\theta}^{(T)}$
\end{algorithmic}
\end{algorithm}

\textbf{Equal Opportunity:} Equalized opportunity requires that prediction $\hat Y$ and the sensitive attribute $A$ are independent conditional on the target $Y = 1$. If the observed true positive rate for a protected class $A = a$ is lower than the expected true positive rate, we increase the weights of the positively labeled samples of $A = a$. This will encourage the classifier to increase its accuracy on the positively labeled samples in $A = a$. In addition, the closer the sample is to the decision boundary in each subgroup, the higher the weight is assigned. Either of these two events will cause the positive prediction rate on $A = a$ to increase and the generalizability of $h_{\theta}$ to improve. This forms the intuition behind Algorithm\ref{alg:eop}.

\begin{algorithm}
\caption{Training a fair classifier for Equal Opportunity}
\label{alg:eop}
\begin{algorithmic}[1]
\State {\bfseries Input:} Training dataset $\mathcal{Z}_{[m]}=\left\{\left(x_i, a_i, y_i\right)\right\}_{i=1}^m$, training iteration $T$, 
subgroup learning rate $\alpha$, step size $\eta$, value of decision boundary $d$, classification procedure $H$.
\State {\bfseries Output:} Fair classifier $h_{\theta}$ for Equal Opportunity.
    \State Initialize the model parameters $\theta$ of the classifier $h$, subgroup weights $W_{1,1}^{(0)}=W_{1,0}^{(0)}=W_{0,1}^{(0)}=W_{0,0}^{(0)}=1$, sample weights $w_1^{(0)}=w_2^{(0)}=\ldots=w_m^{(0)}=\frac{1}{m}$, margin $\phi_1^{(0)}=\phi_2^{(0)}=\ldots=\phi_m^{(0)}=0$, respectively.
    \State Let $h_{\theta}^{(0)}:=H(\mathcal{Z}_{[m]},\{w_i^{(0)}\}_{i=1}^m)$.
    \For{$t$ \textbf{in} $1,\cdots,T$}
        \State Update the margin for each sample:
        \begin{gather*}
            \phi_i^{(t)} = |h_{\theta}^{(t)} - d| \text{ for } i \in [m].
        \end{gather*}
        \For{$a \in \{0, 1\}$}
            \State Calculate subgroup weight $W_{1,a}^{(t)}$:
            \begin{align}
                W_{1,a}^{(t)}=W_{1,a}^{(t-1)} \times \frac{\mid\{i:\hat{y}_{i}^{(t-1)} = 1, y_i = 1\}\mid \times \mid\{i: y_i = 1, a_{i} = a\}\mid + \alpha}{m_{1,\ast} \times \mid\{i: \hat{y}_{i}^{(t-1)} = 1, y_i = 1, a_{i} = a\}\mid + \alpha}.\nonumber
            \end{align}
            \State Update sample weights:
            \begin{gather*}
                w_i^{(t)}=\frac{W_{1,a}^{(t)}}{\sum_{y} \sum_{a} W_{y,a}^{(t)}}\times \frac{p_{1,a}\exp(-\eta \phi_i^{(t)})}{\sum_{i=1}^{m_{1,a}}\exp(-\eta \phi_i^{(t)})} \text{ for } z_i\in G_{1, a},
            \end{gather*}
            \begin{gather*}
                w_i^{(t)}=\frac{1}{m\sum_{y} \sum_{a} W_{y,a}^{(t)}} \text{ for } z_i\in G_{0, a}.
            \end{gather*}
        \EndFor
        \State Update $h_{\theta}^{(t)}:=H(\mathcal{Z}_{[m]},\{w_i^{(t)}\}_{i=1}^m)$.
    \EndFor \\
    \Return $h_{\theta}^{(T)}$
\end{algorithmic}
\end{algorithm}

\subsection{Theoretical Analysis}
In this section, we present theoretical guarantees regarding the learned classifier $h$ when employing the Adaptive Priority Reweighing technique.

\begin{proposition}
    \label{prop1}
    Given the proportion $p_{y, a}$, and considering $wL\left(y, h\left(x, a\right)\right)$ to be upper bounded by $b$, we can state that for any $\delta > 0$, with a probability of at least $1 - \delta$, the following holds:
    \begin{equation}
        \begin{aligned}
            & \sup _{h \in H}\left|R_{L, D}(h)-\hat{R}_{wL, D}(h)\right| \\& =\sup _{h \in H}\left|E_{(X, A, Y) \sim D}\left[\hat{R}_{wL, D}(h)\right]-\hat{R}_{wL, D}(h)\right| \\& \leq \mathop{\max}_{a,y} p_{a,y} \cdot \Re(L \circ H)+2b \sqrt{\frac{\log (1 / \delta)}{2 m}},
        \end{aligned}
    \end{equation}
    where the Rademacher complexity $\Re(L \circ H)$ is defined by~\cite{bartlett2002rademacher}
    \begin{equation}
        \mathfrak{R}(L \circ H)=E_{(X, A, Y) \sim D, \sigma}\left[\sup _{h \in H} \frac{2}{m} \sum_{i=1}^m \sigma_i L\left(y_i, h\left(x_i, a_i\right)\right)\right],
    \end{equation}
    and $\sigma_1, \ldots, \sigma_m$ are i.i.d. Rademacher variables.
\end{proposition}

The Rademacher complexity has a convergence rate of order $\mathcal{O}(\sqrt{1 / n})$~\cite{bartlett2002rademacher}. When the function class satisfies appropriate conditions on its variance, the Rademacher complexity converges rapidly and is of order $\mathcal{O}(1 / n)$~\cite{bartlett2005local}. In Proposition \ref{prop1}, we utilize the Rademacher complexity method to derive the generalization bound. It is worth noting that alternative hypothesis complexities and methods can also be employed to establish the generalization bound.

Considering the inequality:
\begin{equation}
    \begin{aligned}
        & R_{L, D}\left(h'\right)-R_{L, D}\left(h^*\right) \\
        & =R_{wL, D}\left(h'\right)-R_{wL, D}\left(h^*\right) \\
        & \leq 2 \sup _{h \in H}\left|R_{wL, D}(h)-\hat{R}_{wL, D}(h)\right|,
    \end{aligned}
\end{equation}
we observe that the consistency rate is preserved when learning with sample weights.

Building upon Proposition \ref{prop1}, we are now ready to present our main result, which addresses classification in the presence of label bias using the framework of Adaptive Priority Reweighing.

\begin{theorem}
    \label{theorem1}
    The Adaptive Priority Reweighing method allows for the utilization of any surrogate loss functions originally designed for traditional classification problems in the context of classification in the presence of label bias.
\end{theorem}

To derive generalization bounds, we utilize the Rademacher complexity method~\cite{bartlett2002rademacher}.

Let $\sigma_1, \ldots, \sigma_m$ be independent Rademacher variables, and $(x_1,a_1), \ldots, (x_m,a_m)$ be i.i.d. variables. Consider a real-valued function class $H$. The Rademacher complexity of the function class over these variables is defined as follows:
\begin{equation}
\mathfrak{R}(H)=E_{\sigma}\left[\sup _{h \in H} \frac{2}{m} \sum_{i=1}^m \sigma_i h\left(x_i, a_i\right)\right] .
\end{equation}

\begin{theorem}(\cite{bartlett2002rademacher})
    \label{thero2}
    Let $H$ be a real-valued function class on $\mathcal{X} \times \mathcal A$, $S=\left\{(x_1,a_1), \ldots, (x_m,a_m)\right\} \in \mathcal{X}^m \times \mathcal{A}^m$ and
    \begin{equation}
    \Phi(S)=\sup _{h \in H}\left|\frac{1}{m} \sum_{i=1}^m E[h(x, a)]-h\left(x_i, a_i\right)\right|.
    \end{equation}

    Then, $E_S[\Phi(S)] \leq \mathfrak{R}(H)$.
\end{theorem}

The following theorem, which leverages Theorem \ref{thero2} and Hoeffding's inequality, plays a crucial role in establishing the generalization bounds.

\begin{theorem}(\cite{bartlett2002rademacher})
    \label{thero3}
    Let $H$ be an $[a, b]-$valued function class on $\mathcal{X} \times \mathcal A$, and $S=\left\{(x_1,a_1), \ldots, (x_m,a_m)\right\} \in \mathcal{X}^m \times \mathcal{A}^m$. For any $h \in H$ and any $\delta > 0$, there exists a high-probability bound such that, with probability at least $1 - \delta$, the following inequality holds:
    \begin{equation}
    E_{(x,a)}[h(x,a)]-\frac{1}{m} \sum_{i=1}^m h\left(x_i, a_i\right) \leq \mathfrak{R}(H)+(b-a) \sqrt{\frac{\log (1 / \delta)}{2 m}}.
    \end{equation}
\end{theorem}

Based on Theorem \ref{thero3}, it can be readily proven that for any function class with values in the range $[-b, b]$ and any $\delta > 0$, there exists a high-probability bound such that, with probability at least $1 - \delta$, the following statement holds:
\begin{equation}
\begin{aligned}
\sup _{h \in H}\left|E_{(x, a, y) \sim D} \hat{R}_{wL, D}-\hat{R}_{wL, D}\right| \leq \mathfrak{R}(w \circ L \circ H)+2b \sqrt{\frac{\log (1 / \delta)}{2 n}} .
\end{aligned}
\end{equation}

Considering that $w$ is bounded from above by $\mathop{\max}_{a,y} p_{a,y}$, we can utilize the Lipschitz composition property of Rademacher complexity, commonly known as Talagrand's Lemma (refer to, for instance, Lemma 4.2 in~\cite{mohri2018foundations}), to establish the following relationship:
\begin{equation}
\mathfrak{R}(w \circ L \circ H) \leq \mathop{\max}_{a,y} p_{a,y} \mathfrak{R}(L \circ H) .
\end{equation}

Proposition \ref{prop1} can be demonstrated alongside the fact that $E_{(x, a, y) \sim D} \hat{R}{wL, D} = R{wL, D} = R_{L, D}$.

Theorem \ref{theorem1} follows from Proposition \ref{prop1}.

\section{Experiment Details}

This section presents our experiments' implementation details, including the adopted datasets, the investigated fairness-aware algorithms, and experiment settings. 

\subsection{Datasets}

We employ five datasets that have been widely adopted in the context of fairness, including three tabular datasets, one image dataset, and one language dataset:
(1) Adult dataset~\cite{lichman2013uci} consists of $48,842$ samples with gender as its sensitive attribute.
The assigned classification task is to predict whether a person earns over 50K per year.
(2) COMPAS dataset~\cite{angwin2016machine} consists of $7,981$ samples with race as its sensitive attribute.
The assigned task is to predict whether a defendant will re-offend.
(3) Synthetic dataset is generated by CTGAN~\cite{xu2019modeling}, which consists of $70,000$ samples with gender as its sensitive attribute.
The assigned classification task is to predict whether an individual’s income is above $\$ 50,000$.
(4) UTKFace dataset~\cite{zhang2017age} consists of over $20,000$ samples with race as its sensitive attribute.
The assigned classification task is to predict an individual’s gender.
(5) MOJI dataset~\cite{blodgett2016demographic} consists of over $150,000$ samples with race as its sensitive attribute.
The assigned classification task is to predict an individual’s emotions.
For tabular datasets, we follow the same data pre-processing procedures as that in IBM AI Fairness 360~\cite{aif360-oct-2018}, and for vision or language datasets, we follow~\cite{roh2020fairbatch}, or~\cite{ravfogel2020null}.

\begin{table*}
\centering
\caption{Performance on the Adult, COMPAS, and Synthetic test sets w.r.t. equal opportunity. Our method aims to maximize accuracy, denoted by \textuparrow~and to minimize $\Delta_{EOP}$, denoted by \textdownarrow. \textbf{Bold} values represent the best performance.}
\label{tab:adult_EOP}
\begin{tabular}{lcccccc}
\toprule[1pt]
            & \multicolumn{2}{c}{Adult} & \multicolumn{2}{c}{COMPAS} & \multicolumn{2}{c}{Synthetic} \\ \midrule[0.5pt]
Method      & Accuracy↑   & $\Delta_{\mathrm{EOP}}$↓        & Accuracy↑    & $\Delta_{\mathrm{EOP}}$↓        & Accuracy↑   & $\Delta_{\mathrm{EOP}}$↓        \\ \midrule[0.5pt]
LR          & 84.50±0.03  & 9.13±0.03   & 68.58±0.11   & 23.92±0.61 & 83.56±0.02  & 32.31±0.02  \\ \midrule[0.5pt]
Cutting     & 80.22±0.22  & 8.21±1.31   & 67.75±0.64   & 21.51±0.55  & 83.26±0.95  & 30.00±1.21  \\
Reweighing~\cite{kamiran2012data} & 83.41±0.10  & 15.01±0.85  & 68.55±0.08   & 15.70±0.07  & 82.67±0.89  & 8.74±1.01   \\
Fairbatch~\cite{roh2020fairbatch}   & 83.91±0.11  & 2.18±0.33   & 66.13±0.12   & 4.23±0.08   & \textbf{83.19±0.04}  & 0.67±0.31   \\
LBC~\cite{jiang2020identifying}         & \textbf{84.35±0.03}  & 1.19±0.27   & \textbf{68.58±0.05}   & 2.81±0.00   & 82.72±0.07  & 2.33±0.11   \\ \midrule[0.5pt]
FairMixup~\cite{chuang2021fair}  & 81.49±0.12  & 1.01±0.53   & 65.46±0.21   & 2.14±0.35   & 80.54±0.65  & 1.07±0.54   \\
FC~\cite{zafar2017fairness}          & 84.00±0.22  & 2.37±0.85   & 66.68±0.12   & 5.92±1.82   & 81.95±0.61  & 1.46±0.32   \\
AD~\cite{zhang2018mitigating}          & 84.14±0.36  & 4.25±0.66   & 66.39±0.46   & 6.71±2.14   & 80.84±1.11  & 3.51±1.04   \\
RFI~\cite{baharlouei2019renyi}         & 83.86±0.21  & 2.55±0.58   & 66.73±0.22   & 5.18±2.31   & 81.65±0.52  & 1.33±0.41   \\ \midrule[0.5pt]
EO~\cite{hardt2016equality}          & 80.17±0.53  & 4.56±1.35   & 60.31±0.31   & 6.55±1.27   & 77.48±1.40  & 5.18±2.01   \\
CEO~\cite{pleiss2017fairness}         & 79.69±0.45  & 4.56±1.02   & 59.35±0.52   & 7.58±2.03   & 77.35±1.49  & 5.23±1.88   \\ \midrule[0.5pt]
\textbf{Ours}        & 84.34±0.04 & \textbf{0.08±0.03}   & 68.34±0.02 & \textbf{0.06±0.01}   & 82.59±0.00 & \textbf{0.02±0.00}   \\ \bottomrule[1pt]
\end{tabular}
\end{table*}

\begin{table*}
\centering
\caption{Performance on the Adult, COMPAS, and Synthetic test sets w.r.t. equalized odds. Our method aims to maximize accuracy, denoted by \textuparrow~and to minimize $\Delta_{EO}$, denoted by \textdownarrow. \textbf{Bold} values represent the best performance.}
\label{tab:adult_eo}
\begin{tabular}{lcccccc}
\toprule[1pt]
            & \multicolumn{2}{c}{Adult} & \multicolumn{2}{c}{COMPAS} & \multicolumn{2}{c}{Synthetic} \\ \toprule[0.5pt]
Method      & Accuracy↑   & $\Delta_{\mathrm{EO}}$↓         & Accuracy↑   & $\Delta_{\mathrm{EO}}$↓         & Accuracy↑   & $\Delta_{\mathrm{EO}}$↓         \\ \toprule[0.5pt]
LR          & 84.50±0.03  & 9.13±0.03   & 68.58±0.11  & 23.92±0.61  & 83.56±0.02  & 32.31±0.02  \\ \toprule[0.5pt]
Cutting     & 80.22±0.22  & 8.21±1.31   & 67.75±0.64  & 21.51±0.55  & \textbf{83.26±0.95}  & 30.00±1.21  \\
Reweighing~\cite{kamiran2012data} & 83.41±0.10  & 15.01±0.85  & 68.55±0.08  & 15.70±0.07  & 82.67±0.89  & 8.74±1.01   \\
Fairbatch~\cite{roh2020fairbatch}   & 84.21±0.10  & 5.28±0.43   & 68.63±0.12  & 5.21±0.07   & 82.85±0.13  & 1.92±0.10   \\
LBC~\cite{jiang2020identifying}         & 84.30±0.03  & 3.69±0.21   & 68.83±0.05  & 4.16±0.03   & 82.68±0.06  & 1.86±0.11   \\ \toprule[0.5pt]
FairMixup~\cite{chuang2021fair}  & 80.42±0.68  & 5.31±1.36   & 65.43±0.21  & 3.99±0.22   & 80.45±0.22  & 1.18±0.34   \\
FC~\cite{zafar2017fairness}          & 84.10±0.23  & 4.33±0.65   & 67.74±0.16  & 5.92±1.82   & 81.25±0.53  & 1.88±0.42   \\
AD~\cite{zhang2018mitigating}          & \textbf{84.33±0.30}  & 4.35±0.42   & 66.39±0.46  & 6.71±2.14   & 79.88±1.99  & 2.62±0.97   \\
RFI~\cite{baharlouei2019renyi}         & 83.89±0.25  & 4.55±0.53   & 67.73±0.22  & 5.18±2.31   & 80.85±0.50  & 2.01±0.47   \\ \toprule[0.5pt]
EO~\cite{hardt2016equality}          & 80.17±0.53  & 4.56±1.35   & 60.31±0.31  & 6.55±1.27   & 77.38±1.14  & 6.18±1.88   \\
CEO~\cite{pleiss2017fairness}         & 79.69±0.45  & 5.12±1.08   & 59.35±0.52  & 7.58±2.03   & 76.92±1.49  & 6.83±2.13   \\ \toprule[0.5pt]
\textbf{Ours}        & 81.49±0.12 & \textbf{0.87±0.10}   & \textbf{68.99±0.11} & \textbf{0.99±0.15}   & 81.39±0.01 & \textbf{1.09±0.03}   \\ \bottomrule[1pt]
\end{tabular}
\end{table*}

\begin{table*}
\centering
\caption{Performance on the Adult, COMPAS, and Synthetic test sets w.r.t. demographic parity. Our method aims to maximize accuracy, denoted by \textuparrow~and to minimize $\Delta_{DP}$, denoted by \textdownarrow. \textbf{Bold} values represent the best performance.}
\label{tab:adult_dp}
\begin{tabular}{lcccccc}
\toprule[1pt]
            & \multicolumn{2}{c}{Adult} & \multicolumn{2}{c}{COMPAS} & \multicolumn{2}{c}{Synthetice} \\ \toprule[0.5pt]
Method      & Accuracy↑   & $\Delta_{\mathrm{DP}}$↓         & Accuracy↑   & $\Delta_{\mathrm{DP}}$↓         & Accuracy↑   & $\Delta_{\mathrm{DP}}$↓         \\ \toprule[0.5pt]
LR          & 84.50±0.03  & 16.33±0.23  & 68.58±0.11  & 20.45±0.51  & 83.56±0.02  & 26.50±0.02  \\ \toprule[0.5pt]
Cutting     & 80.22±0.22  & 15.22±1.14  & 67.75±0.64  & 9.19±0.53   & \textbf{83.26±0.95}  & 23.70±0.55  \\
Reweighing~\cite{kamiran2012data} & \textbf{83.41±0.10}  & 8.04±0.81   & 68.55±0.08  & 15.77±0.37  & 82.38±0.91  & 13.22±1.20   \\
Fairbatch~\cite{roh2020fairbatch}   & 82.33±0.13  & 1.18±0.13   & 68.13±0.12  & 2.63±0.08   & 81.59±0.07  & 6.03±0.34   \\
LBC~\cite{jiang2020identifying}         & 82.35±0.03  & 1.01±0.07   & \textbf{68.74±0.09}  & 1.27±0.06   & 80.36±0.00  & \textbf{0.06±0.00}   \\ \toprule[0.5pt]
FairMixup~\cite{chuang2021fair}  & 82.46±0.33  & 1.45±0.30   & 64.92±0.41  & 2.10±0.33   & 80.76±0.14  & 0.98±0.31   \\
FC~\cite{zafar2017fairness}          & 82.22±0.22  & 1.45±0.62   & 68.48±0.33  & 5.42±1.62   & 79.52±0.47  & 3.91±0.77   \\
AD~\cite{zhang2018mitigating}          & 81.51±0.76  & 2.64±1.56   & 68.39±0.44  & 6.61±1.18   & 72.39±0.46  & 3.11±1.18   \\
RFI~\cite{baharlouei2019renyi}         & 81.92±0.41  & 1.53±0.56   & 68.66±0.25  & 3.46±1.11   & 79.20±0.28  & 4.83±0.78   \\ \toprule[0.5pt]
EO~\cite{hardt2016equality}          & 80.17±0.53  & 5.31±0.65   & 60.31±0.31  & 5.02±1.26   & 78.47±1.50  & 6.48±0.77   \\
CEO~\cite{pleiss2017fairness}         & 79.69±0.45  & 5.23±0.82   & 59.35±0.52  & 6.08±1.33   & 77.31±1.31  & 6.82±1.13   \\ \toprule[0.5pt]
\textbf{Ours}        & 82.31±0.02 & \textbf{0.14±0.03}   & 68.50±0.04  & \textbf{1.15±0.05}   & 80.38±0.04 & 0.09±0.02   \\ \bottomrule[1pt]
\end{tabular}
\end{table*}

\subsection{Fairness-aware Algorithms}

This subsection investigates ten existing fairness-aware algorithms, including four pre-processing methods, four in-processing methods, and two post-processing methods. For pre-processing methods, we investigate four different approaches: 
(1) \textbf{Cutting}, which equals the data size of different sensitive groups by removing data, (2) \textbf{Reweighing}~\cite{kamiran2012data}, which mitigates discrimination by weighing samples with a fixed weigh, (3) \textbf{Label Bias Correction}(LBC)~\cite{jiang2020identifying}, which iteratively weighs examples towards a fair data distribution, (4) \textbf{Fairbatch}~\cite{roh2020fairbatch}, which adaptively selects minibatch sizes to improve model fairness.
Besides, for in-processing methods, we investigate four different approaches:
(1) \textbf{Fairness Constraints}(FC)~\cite{zafar2017fairness}, which adopts decision boundary covariance as a regularization term, (2) \textbf{Adversarial debiasing}(AD)~\cite{zhang2018mitigating}, which suppresses the dependence between the prediction and sensitive attribute in an adversarial learning manner,
and (3) \textbf{R{\'e}nyi fair inference}(RFI)~\cite{baharlouei2019renyi}, which uses R{\'e}nyi correlation as a regularization term. (4) \textbf{FairMixup}~\cite{chuang2021fair}, which employs mixup paths regularization.
Finally, for post-processing methods, we investigate two different approaches:
(1) \textbf{Equalized Odds} (EO)~\cite{hardt2016equality}, which solves a linear program to optimize equalized odds, (2) \textbf{Calibrated Equalized Odds} (CEO)~\cite{pleiss2017fairness}, which optimizes over calibrated classifier score outputs to find probabilities with which to change output labels with an equalized odds objective.

\subsection{Experiment Settings}
In order to enhance the fairness of the classifier, we employed algorithm \ref{alg:dp}, algorithm \ref{alg:eo}, or algorithm \ref{alg:eop} for training with 200 epochs on the tabular data. Additionally, we performed fine-tuning with 40 epochs for the UTKFace dataset and 35 epochs for the MOJI dataset using the pre-training model. The default batch sizes were set as follows: 1000 for the Adult dataset, 200 for the COMPAS dataset, 2000 for the synthetic dataset, 32 for the UTKFace dataset, and 1024 for the MOJI dataset. In the experiment, we set the learning rate to 0.1, the subgroup learning rate $\alpha$ to a value within the range of [0, 10000], the step size $\eta$ to [0.5, 3], and the decision boundary value $d$ to 0.5. We used the cross-entropy loss function and updated the model parameters using the stochastic gradient descent (SGD) algorithm. Cross-validation was performed on the training set to select the optimal value that resulted in the highest accuracy with minimal fairness violations.

\section{Experiment Analysis}
This section compares our method with the other approaches in the Adult, COMPAS, and synthetic test sets w.r.t. accuracy and three fairness metrics. Furthermore, the experiment shows that our method can improve the fairness of any pre-trained unfair model via fine-tuning.

\subsection{Accuracy and Fairness}
We compare our method with the baseline and three types of fairness-aware algorithms: (1) non-fair method: LR; (2) pre-processing methods: Cutting, Reweighing~\cite{kamiran2012data}, LBC~\cite{jiang2020identifying}, and Fairbatch~\cite{roh2020fairbatch}; (3) in-processing methods: FC~\cite{zafar2017fairness}, AD~\cite{zhang2018mitigating}, RFI~\cite{baharlouei2019renyi}, and FairMixup~\cite{chuang2021fair}; (4) post-processing methods: EO~\cite{hardt2016equality}, and CEO~\cite{pleiss2017fairness}. Table\ref{tab:adult_EOP} compares the performance of our method against the aforementioned SOTA methods on the Adult, COMPAS, and Synthetic test sets w.r.t. equal opportunity. Compared to the logistic regression (LR) method that does not engage with fairness intervention at all, post-processing algorithms (i.e., EO and CEO) improve fairness, but at the same time sacrifice greatly on the accuracy of the classification. In pre-processing methods, Fairbatch and LBC methods adaptively learn the sample probability or weight in each subgroup and are widely acknowledged for performing better than Cutting and Reweighing methods. However, these two methods suffer from poor generalizability, for they assign the same weight to all samples in each subgroup and neglect the differences among samples. Though our method lies in the category of pre-processing methods, we address a well-balanced trade-off between accuracy and fairness. More importantly, our method proves to be more generalizable. Similarly, in the in-processing algorithms, even though FC, AD, and RFI improve algorithmic fairness by training the algorithms with fairness constraints, their performance does not generalize well at the test set. FairMixup attentively addresses the generalizability problem for fairness measures. Sadly, it causes too much damage to the classification accuracy. In comparison, our method not only addresses the generalizability of fair classifiers, we also preserve a more than acceptable level of accuracy. That is, our method performs exceptionally well in improving equal opportunity on the test set without sacrificing much accuracy. To further evaluate our method, we ran experiments on two other most commonly used fairness measures - equalized odds  Table~\ref{tab:adult_eo} and demographic parity Table~\ref{tab:adult_dp}. The results strongly suggest that our method has optimal fairness performance on all three fairness measures across test sets. 

\label{sec:comparison}
\begin{table*}
\centering
\caption{Comparison of model performance on training and test sets in terms of Accuracy and $\Delta_{\text{EOP}}$ on the Adult dataset. Our method aims to maximize accuracy, denoted by \textuparrow~and to minimize $\Delta_{EOP}$, denoted by \textdownarrow.}
\label{tab:train_test_delta_eop}
\begin{tabular}{lcccccc}
\toprule
Method & \multicolumn{2}{c}{Training} & \multicolumn{2}{c}{Test} & \multicolumn{2}{c}{Difference} \\
\cmidrule(r){2-3} \cmidrule(lr){4-5} \cmidrule(l){6-7}
& Accuracy↑ & $\Delta_{\text{EOP}}$↓ & Accuracy↑ & $\Delta_{\text{EOP}}$↓ & Accuracy Diff & $\Delta_{\text{EOP}}$ Diff \\
\midrule
LR        & 84.90±0.02 & 9.32±0.04 & 84.50±0.03 & 9.13±0.03 & -0.40 & -0.19 \\
\midrule
Fairbatch~\cite{roh2020fairbatch} & 84.69±0.03 & 0.37±0.18 & 83.91±0.11 & 2.18±0.33 & -0.78 & 1.81 \\
LBC~\cite{jiang2020identifying}       & 84.40±0.02 & 0.01±0.01 & 84.35±0.03 & 1.19±0.27 & -0.05 & 1.18 \\
\midrule
FairMixup~\cite{chuang2021fair} & 82.43±0.22 & 0.74±0.23 & 81.49±0.12 & 1.01±0.53 & -0.94 & 0.27 \\
FC~\cite{zafar2017fairness}        & 84.56±0.17 & 1.48±0.73 & 84.00±0.22 & 2.37±0.85 & -0.56 & 0.89 \\
RFI~\cite{baharlouei2019renyi}       & 84.06±0.19 & 1.85±0.45 & 83.86±0.21 & 2.55±0.58 & -0.20 & 0.70 \\
\midrule
Ours      & 84.81±0.04 & 0.02±0.02 & 84.34±0.04 & 0.08±0.03 & -0.47 & 0.06 \\
\bottomrule
\end{tabular}
\end{table*}

\begin{table*}
\centering
\caption{Comparison of model performance on training and test sets in terms of Accuracy and $\Delta_{\text{DP}}$ on the Adult dataset. Our method aims to maximize accuracy, denoted by \textuparrow~and to minimize $\Delta_{DP}$, denoted by \textdownarrow.}
\label{tab:train_test_delta_dp}
\begin{tabular}{lcccccc}
\toprule
Method & \multicolumn{2}{c}{Training} & \multicolumn{2}{c}{Test} & \multicolumn{2}{c}{Difference} \\
\cmidrule(r){2-3} \cmidrule(lr){4-5} \cmidrule(l){6-7}
& Accuracy↑ & $\Delta_{\text{DP}}$↓ & Accuracy↑ & $\Delta_{\text{DP}}$↓ & Accuracy Diff & $\Delta_{\text{DP}}$ Diff \\
\midrule
LR        & 84.90±0.02 & 9.32±0.04 & 84.50±0.03 & 9.13±0.03 & -0.40 & -0.19 \\
\midrule
Fairbatch~\cite{roh2020fairbatch} & 83.11±0.17 & 0.39±0.23 & 82.33±0.13 & 1.18±0.13 & -0.78 & 0.79 \\
LBC~\cite{jiang2020identifying}       & 83.00±0.02 & 0.01±0.00 & 82.35±0.03 & 1.01±0.07 & -0.65 & 1.00 \\
\midrule
FairMixup~\cite{chuang2021fair} & 82.61±0.41 & 1.01±0.52 & 82.46±0.33 & 1.45±0.30 & -0.15 & 0.44 \\
FC~\cite{zafar2017fairness}        & 82.48±0.21 & 0.89±0.50 & 82.22±0.22 & 1.45±0.62 & -0.26 & 0.56 \\
RFI~\cite{baharlouei2019renyi}       & 81.88±0.53 & 1.32±0.47 & 81.92±0.41 & 1.53±0.56 & 0.04 & 0.21 \\
\midrule
Ours      & 82.69±0.08 & 0.02±0.01 & 82.31±0.02 & 0.14±0.03 & -0.38 & 0.12 \\
\bottomrule
\end{tabular}
\end{table*}

\begin{table*}
\centering
\caption{Comparison of model performance on training and test sets in terms of Accuracy and $\Delta_{\text{EO}}$ on the Adult dataset. Our method aims to maximize accuracy, denoted by \textuparrow~and to minimize $\Delta_{EO}$, denoted by \textdownarrow.}
\label{tab:train_test_delta_eo}
\begin{tabular}{lcccccc}
\toprule
Method & \multicolumn{2}{c}{Training} & \multicolumn{2}{c}{Test} & \multicolumn{2}{c}{Difference} \\
\cmidrule(r){2-3} \cmidrule(lr){4-5} \cmidrule(l){6-7}
& Accuracy↑ & $\Delta_{\text{EO}}$↓ & Accuracy↑ & $\Delta_{\text{EO}}$↓ & Accuracy Diff & $\Delta_{\text{EO}}$ Diff \\
\midrule
LR        & 84.90±0.02 & 9.32±0.04 & 84.50±0.03 & 9.13±0.03 & -0.40 & -0.19 \\
\midrule
Fairbatch~\cite{roh2020fairbatch} & 84.72±0.12 & 4.54±0.22 & 84.21±0.10 & 5.28±0.43 & -0.51 & 0.74 \\
LBC~\cite{jiang2020identifying}       & 84.55±0.07 & 3.44±0.06 & 84.30±0.03 & 3.69±0.21 & -0.25 & 0.25 \\
\midrule
FairMixup~\cite{chuang2021fair} & 81.09±1.48 & 5.44±1.08 & 80.42±0.68 & 5.31±1.36 & -0.67 & -0.13 \\
FC~\cite{zafar2017fairness}        & 84.52±0.18 & 3.82±0.72 & 84.10±0.23 & 4.33±0.65 & -0.42 & 0.51 \\
RFI~\cite{baharlouei2019renyi}       & 84.22±0.31 & 4.60±0.46 & 83.89±0.25 & 4.55±0.53 & -0.33 & -0.05 \\
\midrule
Ours      & 81.66±0.16 & 0.89±0.12 & 81.49±0.12 & 0.87±0.10 & -0.17 & -0.02 \\
\bottomrule
\end{tabular}
\end{table*}

\subsubsection{Comparison of Model Performance on Training and Test Sets}
\label{611}

The discrepancy in model performance, referred to as 'diff,' is calculated by subtracting the performance metrics obtained on training sets from those on test sets. This discrepancy highlights the challenge of generalizing fairness measures, an issue that becomes pronounced in algorithms lauded for their fairness, such as LBC and Fairbatch. For instance, while these algorithms exhibit commendable fairness on training sets—particularly in terms of the equal opportunity metric—their fairness does not consistently extend to the test sets.

Our comparative analysis (using the Adult dataset as an example), focusing on Equal Opportunity (refer to Table~\ref{tab:train_test_delta_eop}) and Demographic Parity (refer to Table~\ref{tab:train_test_delta_dp}), reveals that our method excels in fairness generalizability, as also illustrated in Figure~\ref{fig:compare}. When considering Equalized Odds (see Figure~\ref{tab:train_test_delta_eo}), we find no significant generalizability concerns among the six evaluated methods. This could be because none of the methods particularly excel in ensuring fairness, as measured by Equalized Odds, even on the training set.

Fairmixup closely follows our method in terms of fairness generalizability. This can be attributed to its inclusion of constraint terms in its algorithm, specifically addressing the critical issue of fairness generalization.

\subsubsection{The trade-offs between accuracy and fairness}

\begin{figure}
    \centering
    \includegraphics[width=1\linewidth]{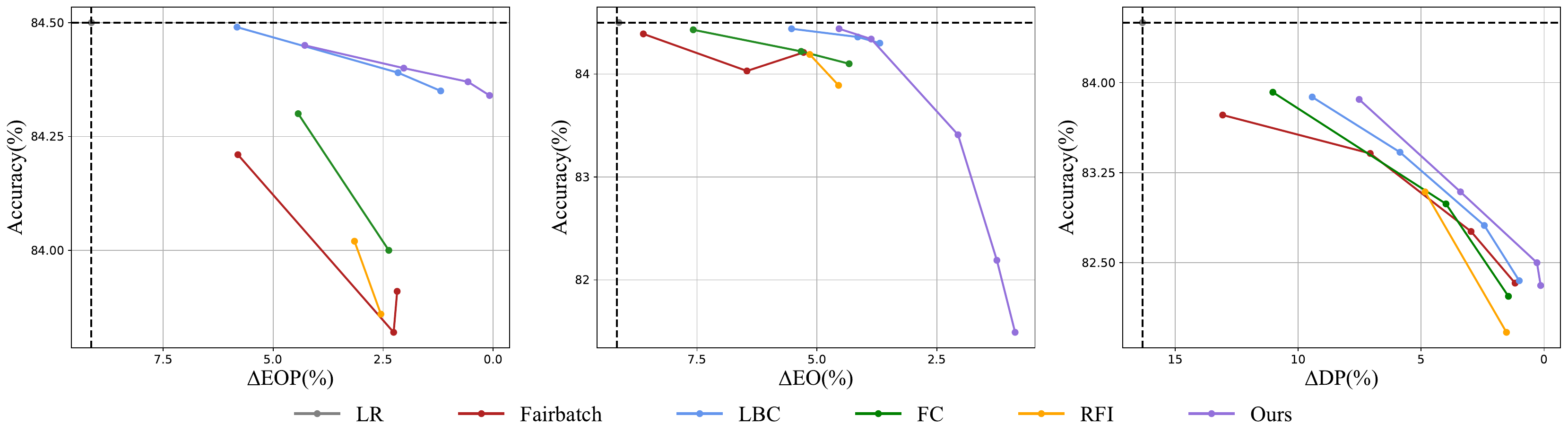}
    \caption{The trade-offs between accuracy and fairness measures on Adult. The dashed line represents the accuracy and fairness metrics corresponding to the baseline (LR). The upper-right corner of each diagram indicates optimal performance for both accuracy and fairness.}
\label{fig:trade-off}
\end{figure}

As depicted in Figure \ref{fig:trade-off}, We selected four other algorithms with notable fairness performance and conducted a comparative analysis of the trade-offs between accuracy and three fairness metrics on the Adult datasets. Our method demonstrates superior performance in balancing the trade-offs between accuracy and fairness in both Equal Opportunity and Demographic Parity metrics. In terms of the Equalized Odds metric, as illustrated, our method sacrifices a substantial amount of accuracy when attempting to enhance fairness. We highlight the trade-off effects between accuracy and fairness, calling for future research on algorithmic fairness to investigate these dynamics, especially on the fairness metric - Equalized Odds.

\subsubsection{Case Study}

\begin{table*}
\centering
\caption{A selective case analysis on the Adult dataset. We examine the classification of income groups based on individual attributes. The analysis was specifically centered around cases where Income>50K, employing the Equal Opportunity metric as the basis for the analysis.}

\label{tab:detailed_cases}
\begin{tabular}{l|c|c|c|c}
\hline
Attributes & Case 1 & Case 2 & Case 3 & Case 4 \\
\hline
Age & 49 & 40 & 45 & 47 \\
Workclass & Private & Private & Private & Self-emp-not-inc \\
Education & HS-grad & HS-grad & Bachelors & 5th-6th \\
Education-Num & 9 & 9 & 13 & 3 \\
Marital-Status & Married-civ-spouse & Married-civ-spouse & Divorced & Married-civ-spouse \\
Occupation & Craft-repair & Exec-managerial & Other-service & Sales \\
Relationship & Wife & Husband & Not-in-family & Husband \\
Race & White & White & White & White \\
Sex & Female & Male & Male & Male \\
Capital-Gain & 0 & 0 & 0 & 0 \\
Capital-Loss & 0 & 0 & 0 & 0 \\
Hours-Per-Week & 40 & 40 & 45 & 55 \\
Native-Country & United-States & United-States & ? & Italy \\
\hline
Income & >50K & >50K & >50K & >50K \\
Prediction & <=50K & <=50K & <=50K & <=50K \\
\hline
\end{tabular}
\end{table*}

We conducted further investigation into instances of misclassification, delving into the reasons persisting despite the implementation of our method (Table \ref{tab:detailed_cases}). We present specific cases to demonstrate that our method achieves the purpose of debias by providing alternative explanations for the misclassification instances, highlighting that bias is not the root cause.

\begin{itemize}
    \item Case 1: The individual's education level (HS-Grad) and Occupation (Craft repair) are generally not directly associated with high income, which may contribute to the error of classifying it as "<=50K". Gender is usually a sensitive feature closely associated with high income, in this case, however, its impact on model prediction might be relatively minor compared to that of education and occupation.
    \item Case 2: Though the individual's occupation is Exec-managerial, which is typically associated with higher income levels, it may not be typical for individuals in this occupation with an HS-grad education level to have higher income. This could result in the model incorrectly predicting their income as "<=50K", especially when high-income individuals in the dataset typically have higher education levels.
    \item Case 3: Despite being a 45-year-old male with a bachelor's degree, which is usually a positive indicator of higher income, his occupation is listed as "other-service". which is generally considered a low-paying job. In addition, the model is likely to associate a "divorced" marital status and "not-in-family" with lower income, especially when there lacks strong indicators supporting a high-income prediction. Also, the absence of nationality information may prove to be challenging for the model to accurately assess his income, especially when certain nationalities are positively associated with specific economic opportunities.
    \item Case 4: Despite working in sales and potentially having higher-income potential as a self-employed individual, this 47-year-old male has a lower education level (5th-6th grade), which typically does not align with a high-income group, leading to a potential underestimation of his income ability. Additionally, this individual works in Italy which indicates certain market and economic opportunities, nevertheless, the model may fail to fully understand income potential in different national contexts, especially for self-employed individuals.
\end{itemize}

\subsection{Fine-Tuning Pretrained Unfair Models for Fairness}

\begin{table}
\centering
\caption{Fine-tuning the Performance of pre-trained models with our method. Our method aims to maximize accuracy, denoted by \textuparrow~and to minimize $\Delta_{EO}$, denoted by \textdownarrow. \textbf{Bold} values represent the best performance.}
\label{tab:pretrain}
\begin{tabular}{lllcc}
\toprule[1pt]
Dataset                  & \begin{tabular}[c]{@{}l@{}}Pre-trained\\   model\end{tabular} & Method      & Accuracy↑  & $\Delta_{\mathrm{EO}}$↓        \\ \midrule[0.5pt]
\multirow{6}{*}{UTKFace} & \multirow{6}{*}{ResNet18}                                     & Original    & 89.13±0.61 & 9.12±1.01  \\ 
                         &                                                               & Reweighing~\cite{kamiran2012data}  & 87.70±0.73 & 8.06±0.82  \\
                         &                                                               & Fairbatch~\cite{roh2020fairbatch}   & 89.08±0.20 & 6.46±0.18  \\
                         &                                                               & LBC~\cite{jiang2020identifying}         & \textbf{89.11±0.13} & 6.03±0.10  \\
                         &                                                               & \textbf{Ours}        & 89.06±0.55 & \textbf{5.88±0.64}  \\ \midrule[0.5pt]
\multirow{7}{*}{MOJI}    & \multirow{7}{*}{DeepMoji}                                     & Original    & 72.11±0.11 & 28.56±0.41 \\
                         &                                                               & Reweighing~\cite{kamiran2012data}  & 73.29±0.36 & 18.03±0.39 \\
                         &                                                               & FairBatch~\cite{roh2020fairbatch}   & \textbf{75.03±0.33} & 10.22±0.61 \\
                         &                                                               & LBC~\cite{jiang2020identifying}         & 74.99±0.21 & 8.89±0.30  \\
                         &                                                               & Gate~\cite{han2022balancing}        & 74.55±0.23 & 8.63±0.51  \\
                         &                                                               & \textbf{Ours}        & 74.59±1.03 & \textbf{8.02±1.83}  \\
\bottomrule[1pt]
\end{tabular}
\end{table}

Having evaluated our method against state-of-the-art algorithms on tabular benchmarks (Table \ref{tab:adult_EOP}, Table \ref{tab:adult_eo}, and Table \ref{tab:adult_dp}), we extended our analysis to vision and language data, employing widely recognized models such as ResNet18~\cite{he2016deep} and DeepMoji~\cite{felbo2017using}. These models, despite their effectiveness, have raised fairness concerns in their respective domains\cite{roh2020fairbatch,ravfogel2020null}, motivating us to explore fine-tuning techniques to enhance fairness while maintaining model accuracy.

In the comprehensive comparison detailed in Table \ref{tab:pretrain}, we juxtaposed our method with the Reweighing approach, Fairbatch~\cite{roh2020fairbatch}, and LBC~\cite{jiang2020identifying}—methods previously validated in tabular contexts but here applied to vision and language tasks. Among these, our method distinguished itself by not just matching the accuracy levels of these established methods but also by leading in fairness metrics, achieving the lowest $\Delta_{\mathrm{EO}}$ across the board. This singular achievement highlights the capability of our approach to enhance fairness effectively without detracting from the predictive performance of the models. We also explored the Gate~\cite{han2022balancing} method, notable for its application of reweighting and auxiliary networks to improve fairness in language processing, which demonstrated notable improvements as well.

The collective results from our evaluation illuminate the diverse landscape of fairness enhancement methods, each contributing to the reduction of fairness disparities in their own right. Among these, our approach stands out by striking a nuanced balance between fairness and accuracy, which is pivotal in nuanced and multifaceted contexts like AI. This balance not only underscores the adaptability of our method across various data types—from tabular to the more complex vision and language tasks—but also highlights its potential to thoughtfully address fairness concerns within these domains.

\subsection{Evaluating Impacts of \texorpdfstring{$\alpha$ and $\eta$}{alpha and eta} Parameters}

\begin{figure}
    \centering
    \includegraphics[width=1\linewidth]{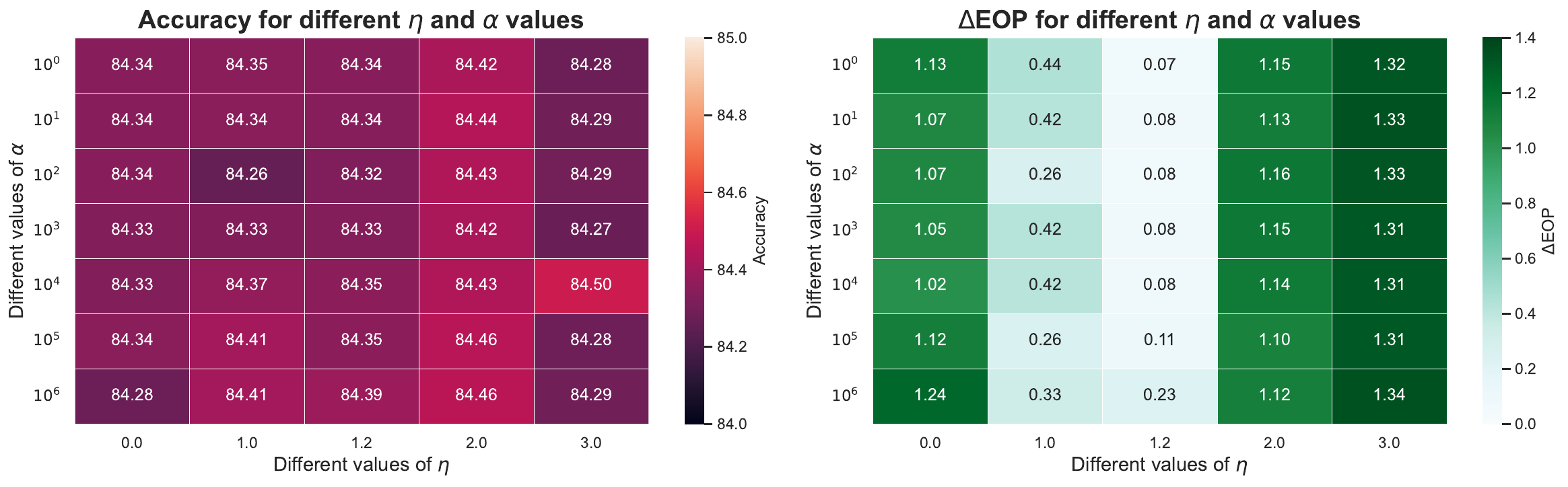}
    \caption{Results obtained for different $\alpha$ and different $\eta$ on the Adult test set w.r.t. equal opportunity.}
\label{fig:adult_EOP}
\end{figure}

\begin{figure}
    \centering
    \includegraphics[width=1\linewidth]{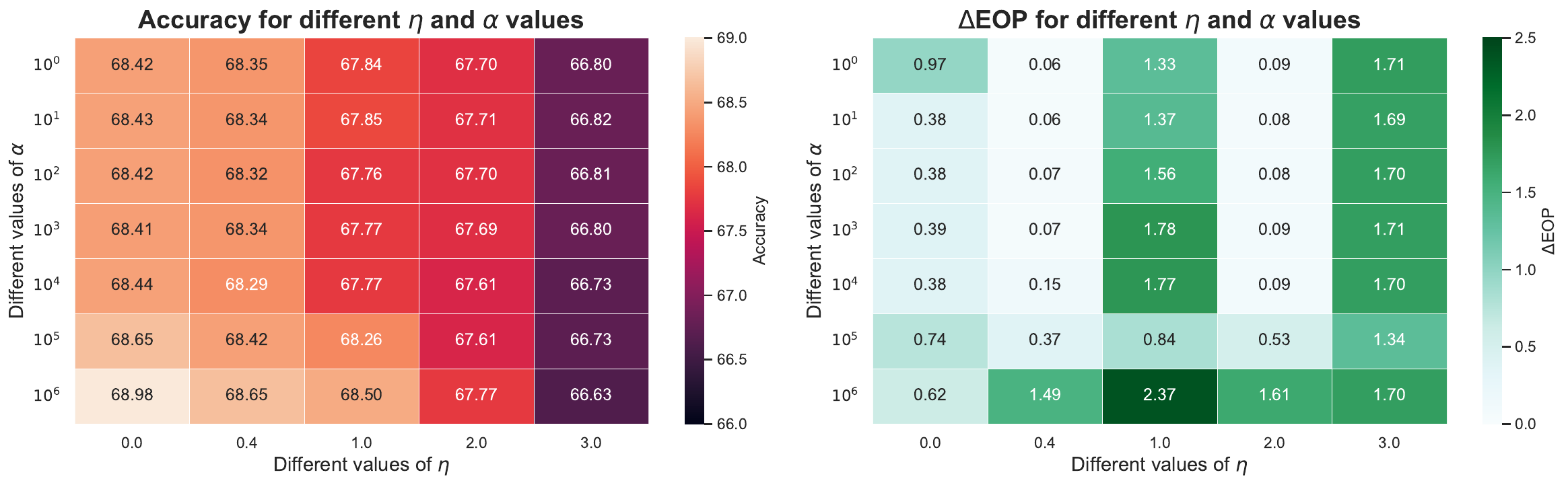}
    \caption{Results obtained for different $\alpha$ and different $\eta$ on the COMPAS test set w.r.t. equal opportunity.}
\label{fig:compas_EOP}
\end{figure}

\begin{figure}
    \centering
    \includegraphics[width=1\linewidth]{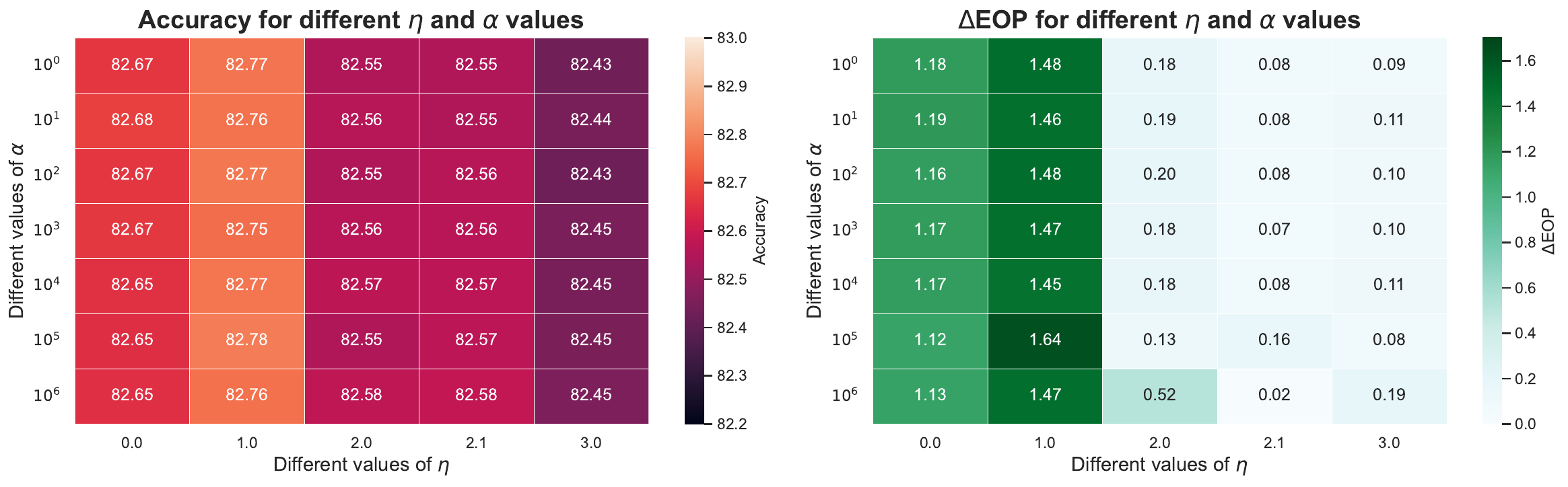}
    \caption{Results obtained for different $\alpha$ and different $\eta$ on the Synthetic test set w.r.t. equal opportunity.}
\label{fig:syn_EOP}
\end{figure}

\begin{figure}
    \centering
    \includegraphics[width=1\linewidth]{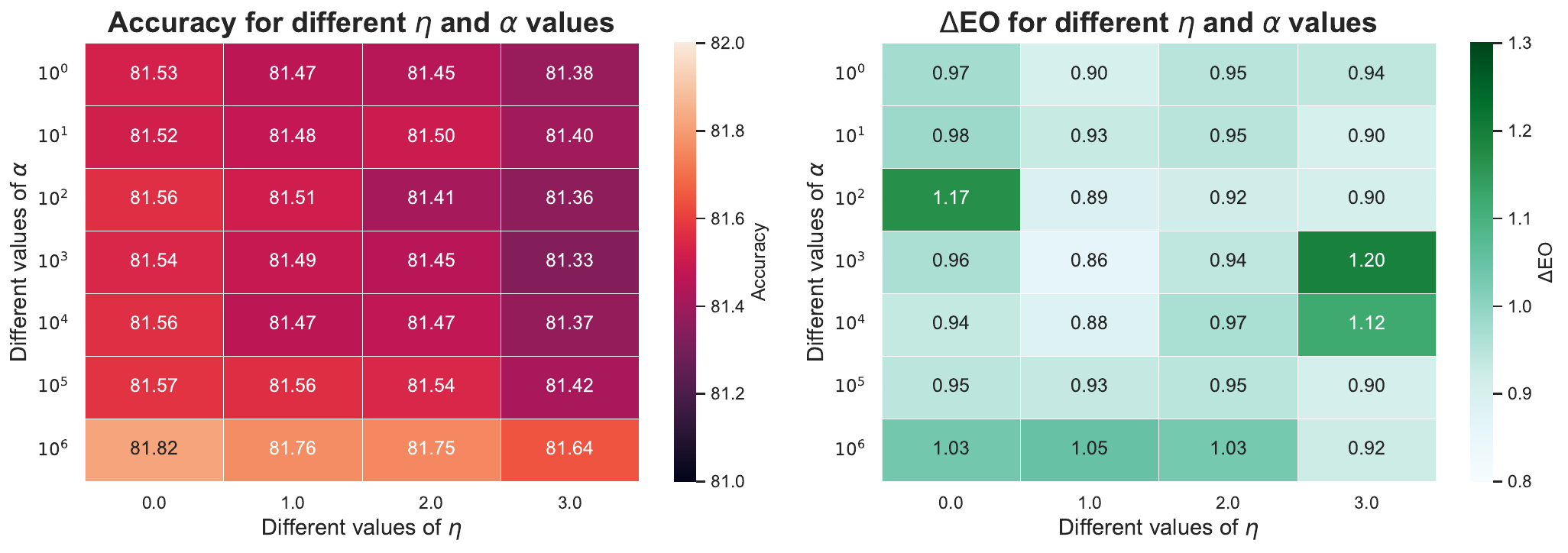}
    \caption{Results obtained for different $\alpha$ and different $\eta$ on the Adult test set w.r.t. equalized odds.}
\label{fig:adult_eo}
\end{figure}

\begin{figure}
    \centering
    \includegraphics[width=1\linewidth]{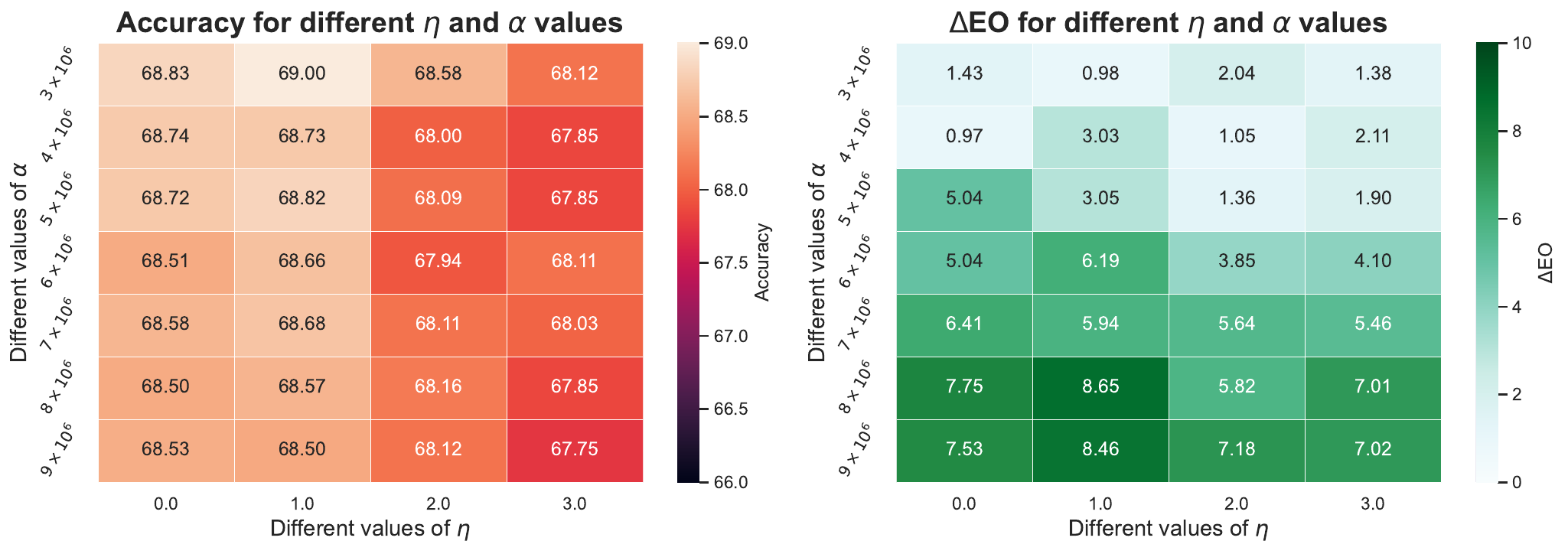}
    \caption{Results obtained for different $\alpha$ and different $\eta$ on the COMPAS test set w.r.t. equalized odds.}
\label{fig:compas_eo}
\end{figure}

\begin{figure}
    \centering
    \includegraphics[width=1\linewidth]{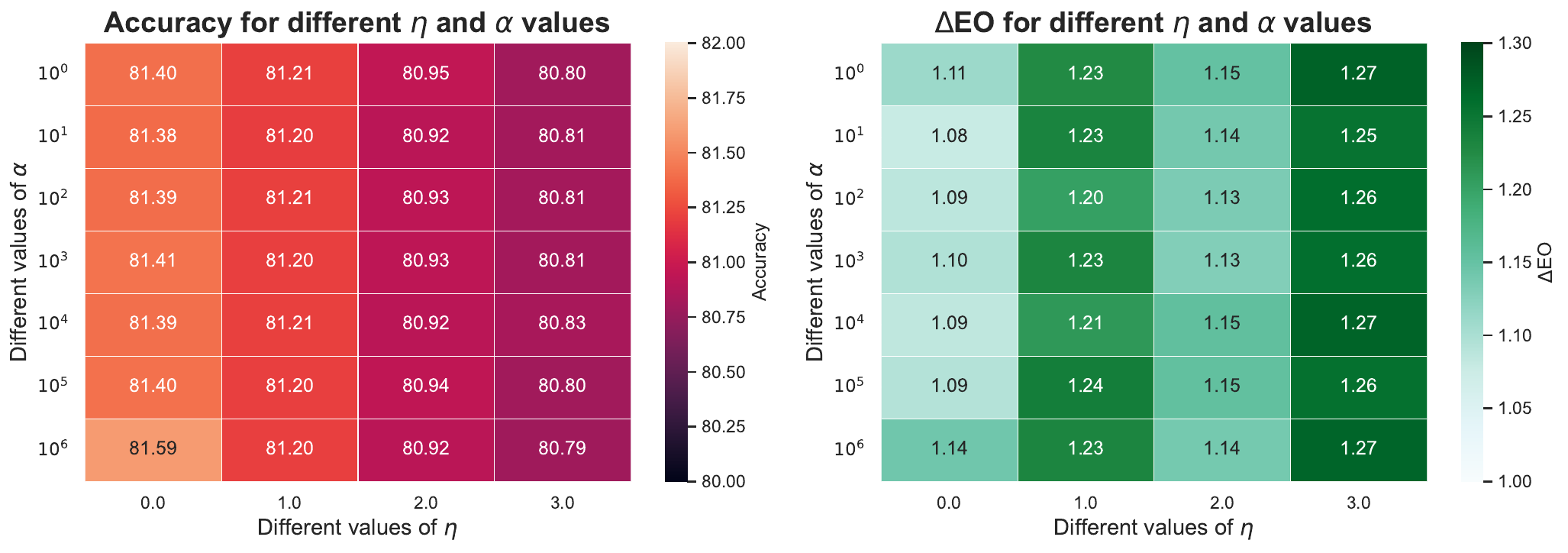}
    \caption{Results obtained for different $\alpha$ and different $\eta$ on the Synthetic test set w.r.t. equalized odds.}
\label{fig:syn_eo}
\end{figure}

\begin{figure}
    \centering
    \includegraphics[width=1\linewidth]{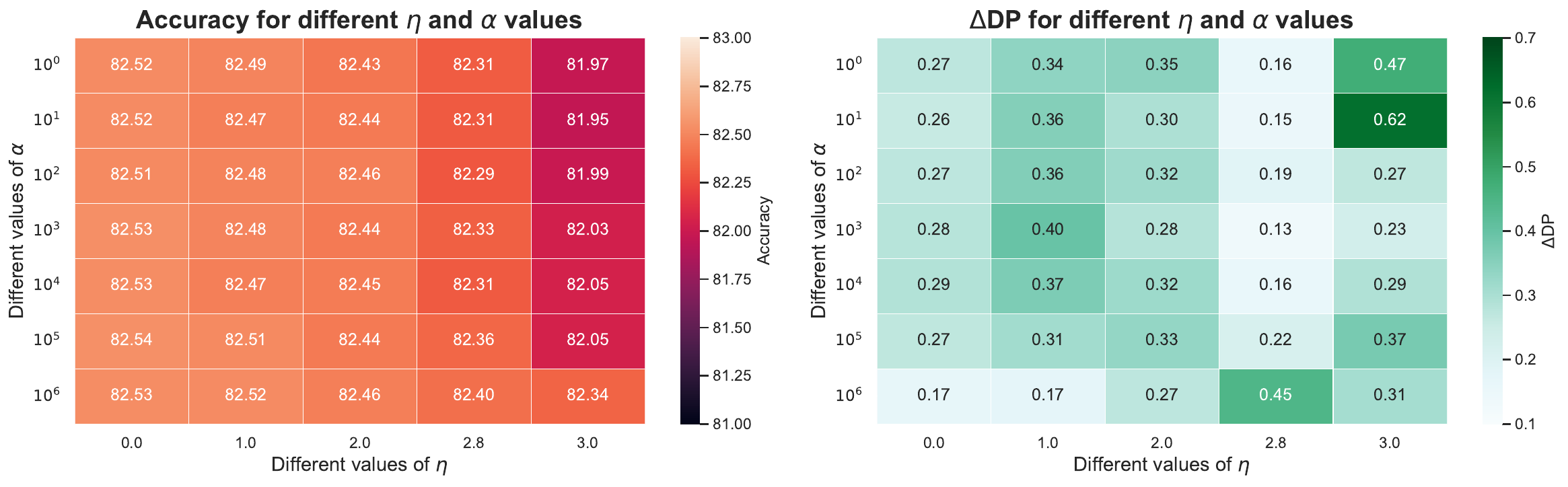}
    \caption{Results obtained for different $\alpha$ and different $\eta$ on the Adult test set w.r.t. demographic parity.}
\label{fig:adult_dp}
\end{figure}

\begin{figure}
    \centering
    \includegraphics[width=1\linewidth]{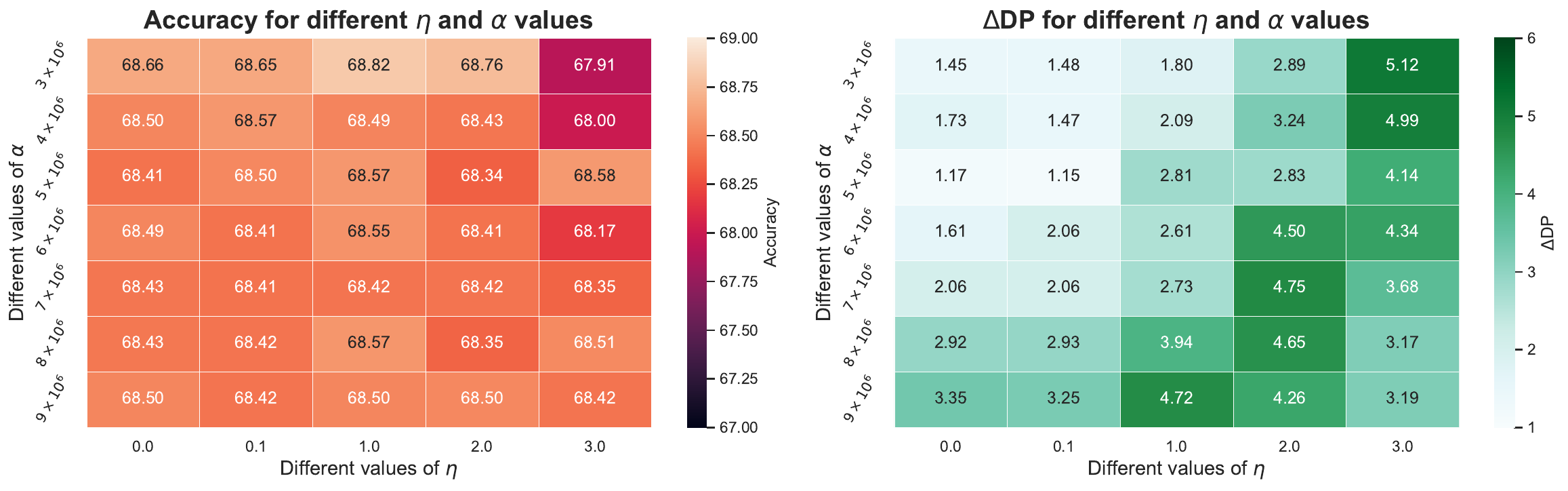}
    \caption{Results obtained for different $\alpha$ and different $\eta$ on the COMPAS test set w.r.t. demographic parity.}
\label{fig:compas_dp}
\end{figure}

\begin{figure}
    \centering
    \includegraphics[width=1\linewidth]{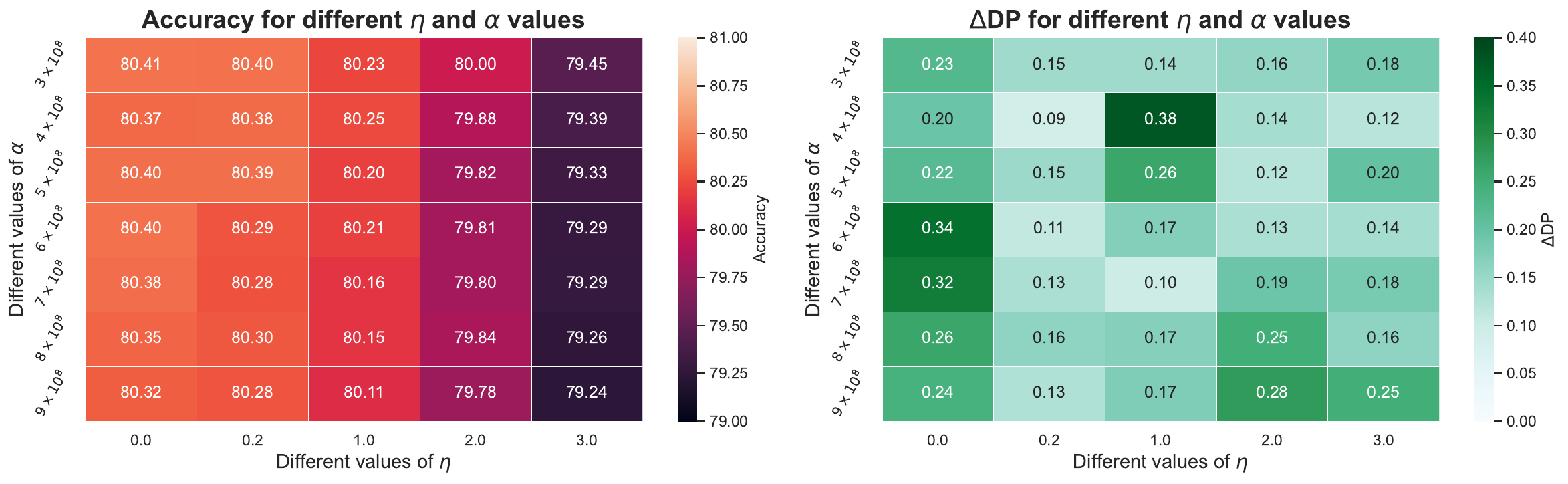}
    \caption{Results obtained for different $\alpha$ and different $\eta$ on the Synthetic test set w.r.t. demographic parity.}
\label{fig:syn_dp}
\end{figure}

We examine the performance of our adaptive priority reweighing method for accuracy and three fairness measures - equal opportunity (Figures~\ref{fig:adult_EOP},~\ref{fig:compas_EOP},~\ref{fig:syn_EOP}), equalized odds (Figures~\ref{fig:adult_eo},~\ref{fig:compas_eo},~\ref{fig:syn_eo}), and demographic parity (Figures~\ref{fig:adult_dp},~\ref{fig:compas_dp},~\ref{fig:syn_dp} across three datasets - Adult, COMPAS, and Synthetic dataset.
We illustrate the performance using color - the lighter the shade indicates better performance on accuracy and fairness. The adaptive priority reweighing algorithm is governed by two pivotal hyperparameters: the subgroup learning rate, $\alpha$, and the step size, $\eta$. In this section, we delve into the effects of varying these hyperparameters on the algorithm's performance.

From these figures, we validated that applying adjustment to $\eta$  
and prioritizing those closer to the decision boundary with higher weight improves the generalizability of fair classifiers. More specifically, when $\eta$>0, our method proves to optimize fairness on the equal opportunity measure (Figures~\ref{fig:adult_EOP},~\ref{fig:compas_EOP},~\ref{fig:syn_EOP}) as well as on the demographic parity measure (Figures~\ref{fig:adult_dp},~\ref{fig:compas_dp},~\ref{fig:syn_dp})
across all datasets. Though the performance of our model on improving fairness in accordance with the equalized odds measure was not evident, especially on the Synthetic dataset, it corresponds with the premises of our study that our method enables fairness improvement to fairness measures on the test dataset when they have already performed well on the training dataset. The equalized odds measure (see Table~\ref{tab:train_test_delta_eo} and Figure~\ref{fig:trade-off}) shows limitations on the training set, as we specified in section~\ref{611}.

However, as illustrated in the figures, owing to the fact that increasing $\eta$ allocates higher weight to samples closer to the decision boundary, augmenting it excessively can result in a decline in both accuracy and fairness performance.  Specifically, in the case of equal opportunity, adjusting $\alpha$ shows minimal impact on both accuracy and fairness, whereas $\eta$, under appropriate settings, enhances model's fairness performance on the test set substantially. Similarly, in cases of equalized odds and demographic parity, $\eta$ holds a greater influence on accuracy and fairness. However, adjusting $\alpha$ within these fairness measures exhibits a more noticeable impact on fairness.

A cursory observation across all tables indicates that the performance metrics exhibit limited variation with increasing values of $\alpha$, suggesting a relatively subdued influence of $\alpha$ on the algorithm. Conversely, the step size, $\eta$, appears to exert a more pronounced effect. For instance, Figure \ref{fig:adult_EOP} reveals that optimal fairness is predominantly achieved at $\eta \approx 1.2$ for the Adult dataset. In the case of the COMPAS dataset, a consistent $\eta$ of 0.4 yields optimal fairness across a spectrum of $\alpha$ values. This behavior underscores that $\alpha$ primarily serves to cushion abrupt shifts during weight iteration updates, while adjusting $\eta$ accentuates the weights of samples near the decision boundary, bolstering fairness generalization.

\begin{table*}
\centering
\caption{Performance comparison of different baselines with our method. Our method aims to maximize accuracy, denoted by \textuparrow~and to minimize $\Delta_{EO}$, denoted by \textdownarrow.}
\label{tab:baseline_comparison}
\begin{tabular}{lllcc}
\toprule[1pt]
Dataset & Baselines & Method & Accuracy↑ & $\Delta_{\mathrm{EO}}$↓ \\
\midrule[0.5pt]
\multirow{6}{*}{Adult} & \multirow{2}{*}{LR} & Original & 84.50±0.03 & 9.13±0.03 \\
                       &                     & Ours    & 81.49±0.12 & 0.87±0.10 \\
\cmidrule{2-5}
                       & \multirow{2}{*}{GBDT} & Original & 86.53±0.11 & 8.73±0.09 \\
                       &                      & Ours    & 84.51±0.13 & 0.85±0.17 \\
\cmidrule{2-5}
                       & \multirow{2}{*}{XGBoost} & Original & 86.53±0.04 & 8.11±0.06 \\
                       &                        & Ours    & 84.03±0.07 & 0.82±0.43 \\
\midrule[0.5pt]
\multirow{4}{*}{UTKFace} & \multirow{2}{*}{ResNet18} & Original & 89.13±0.61 & 9.12±1.01 \\
                         &                           & Ours    & 89.06±0.55 & 5.88±0.64 \\
\cmidrule{2-5}
                         & \multirow{2}{*}{ResNet50} & Original & 89.25±0.43 & 8.53±0.88 \\
                         &                           & Ours    & 89.14±0.36 & 5.92±0.61 \\
\midrule[0.5pt]
\multirow{4}{*}{MOJI}    & \multirow{2}{*}{DeepMoji} & Original & 72.11±0.11 & 28.56±0.41 \\
                         &                           & Ours    & 74.59±1.03 & 8.02±1.83 \\
\cmidrule{2-5}
                         & \multirow{2}{*}{Bert}     & Original & 72.85±0.18 & 31.01±0.60 \\
                         &                           & Ours    & 75.01±1.22 & 7.95±1.46 \\
\bottomrule[1pt]
\end{tabular}
\end{table*}

Certain $\alpha$ and $\eta$ combinations yield optimal fairness for specific datasets. A case in point is the Synthetic dataset in Figure\ref{fig:syn_dp}, where an $\alpha$ of $4 \times 10^8$ coupled with $\eta$ at 0.2 minimizes $\Delta_{\mathrm{DP}}$ to a value of 0.09. A noteworthy observation is the heightened sensitivity of the COMPAS dataset to alterations in both hyperparameters, manifesting in substantial variations in its performance metrics compared to the Adult and Synthetic datasets. The intricacies in the interplay between $\alpha$ and $\eta$ and their resultant effect on the metrics across datasets emphasize the nuanced relationship between these parameters and algorithmic performance. Certain parameter configurations manage to strike a harmonious balance, ensuring both high accuracy and commendable fairness.

\subsection{Performance Comparison of Different Baselines}
The experiments aim to validate the efficacy of our method in improving model fairness across various baselines (Table~\ref{tab:baseline_comparison}). In addition to the conventional models such as LR, ResNet18, and DeepMoji, we further explored higher-performing baselines in tabular (i.e., GBDT~\cite{friedman2001greedy} and XGBoost~\cite{chen2016xgboost}), vision (i.e., ResNet50~\cite{he2016deep}), and language (i.e., Bert~\cite{kenton2019bert}) dataset. Leveraging these more advanced baselines, our method consistently achieves improvement in classification accuracy while maintaining fairness performance. Especially on tree models such as GBDT and XGBoost, our method demonstrates significant improvements compared to the original models. Maintaining equivalent fairness performance, our method outperforms LR by a notable three percent.

\section{Conclusion}
This paper responds to the calls for fair algorithms that generalize well across the test set by designing a novel reweighing method – the adaptive priority reweighing method. This method eliminates the impact of distribution shifts between training and testing data on model generalizability, a conundrum constantly faced with equal reweighing methods.

The performance of our method for accuracy and fairness measures (i.e., equal opportunity, equalized odds, and demographic parity) is evaluated through extensive experiments on tabular datasets. We further highlight the performance of our method in improving the generalizability of fair classifiers by experimenting with both language and vision benchmarks. We believe that our method performs extremely well in improving fairness and will show promising results in improving the fairness of any pre-trained models simply via fine-tuning.

\bibliographystyle{ACM-Reference-Format}
\bibliography{references.bib}
\end{document}